\documentclass[lettersize,journal]{IEEEtran}
\usepackage{amsmath,amsfonts}
\usepackage{url}
\usepackage{booktabs} 
\usepackage{bm}
\usepackage{enumitem}
\usepackage{multirow}
\usepackage{colortbl}
\usepackage{algorithm}
\usepackage{algpseudocode}
\usepackage{setspace}
\usepackage{makecell}
\usepackage{bbding}
\usepackage{soul}
\usepackage{xcolor}
\usepackage{mathtools}

\def\colorfirst{\cellcolor[HTML]{9698ED}}
\def\colorsecond{\cellcolor[HTML]{CBCEFB}}
\def\colorthird{\cellcolor[HTML]{ECF4FF}}
\def\squarefirst{\textcolor[HTML]{9698ED}{\rule{2mm}{2mm}}}
\def\squaresecond{{\textcolor[HTML]{CBCEFB}{\rule{2mm}{2mm}}}}
\def\squarethird{{\textcolor[HTML]{ECF4FF}{\rule{2mm}{2mm}}}}

\usepackage{graphicx}
\usepackage{etoolbox}
\usepackage{caption}
\newcommand{\insertfig}{\setcounter{figure}{0}\includegraphics[width=0.95\linewidth]{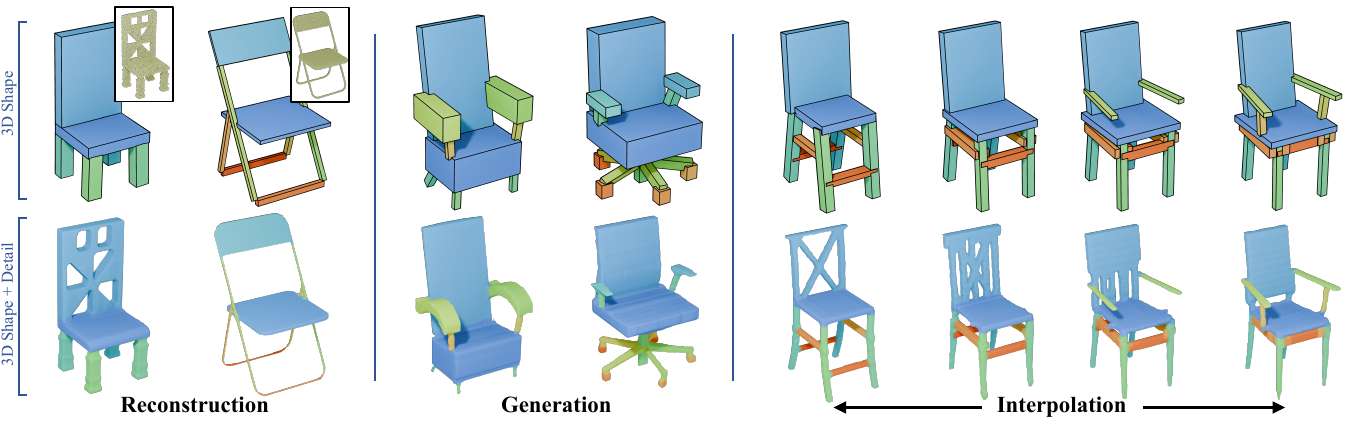}\captionof{figure}{We parameterize the structures of 3D shapes through differentiable templates and utilize the three-view details to represent their inside details. Here, the results of our method show that our method can reconstruct and generate diverse shapes with complicated details, and interpolate them smoothly.}\label{fig:teaser}}

\makeatletter
\apptocmd{\@maketitle}{\centering\insertfig}{}{}
\makeatother

\begin{document}

\title{Parameterize Structure with Differentiable Template for 3D Shape Generation}

\author{Changfeng Ma, Pengxiao Guo, Shuangyu Yang, Yinuo Chen, Jie Guo, Chongjun Wang, Yanwen Guo, and Wenping Wang,~\IEEEmembership{Fellow,~IEEE}
\IEEEcompsocitemizethanks{\IEEEcompsocthanksitem C. Ma, P. Guo, S. Yang, Y. Chen, J. Guo, C. Wang, Y. Guo are with the National Key Lab for Novel Software Technology, Nanjing University, Nanjing 210000, China ( e-mail: changfengma @ smail.nju.edu.com; px guo @ smail.nju.edu.cn; shuangyuyang @ smail.nju.edu.cn; yinuochen @ smail.nju.edu.cn ; guojie @ nju.edu.cn; chjwang @ nju.edu.cn; ywguo @ nju.edu.cn). 
\IEEEcompsocthanksitem W. Wang is with the Texas A\&M University, United States of America (e-mail: wenping@cs.hku.hk).
\IEEEcompsocthanksitem (Corresponding author: Y. Guo.)
}
}
\markboth{Journal of \LaTeX\ Class Files,~Vol.~14, No.~8, August~2021}%
{Shell \MakeLowercase{\textit{et al.}}: A Sample Article Using IEEEtran.cls for IEEE Journals}


\maketitle

\begin{abstract}
Structural representation is crucial for reconstructing and generating editable 3D shapes with part semantics. 
Recent 3D shape generation works employ complicated networks and structure definitions relying on hierarchical annotations and pay less attention to the details inside parts.
In this paper, we propose the method that parameterizes the shared structure in the same category using a differentiable template and corresponding fixed-length parameters.
Specific parameters are fed into the template to calculate cuboids that indicate a concrete shape. 
We utilize the boundaries of three-view drawings of each cuboid to further describe the inside details.
Shapes are represented with the parameters and three-view details inside cuboids, from which the SDF can be calculated to recover the object.
Benefiting from our fixed-length parameters and three-view details, our networks for reconstruction and generation are simple and effective to learn the latent space.
Our method can reconstruct or generate diverse shapes with complicated details, and interpolate them smoothly.
Extensive evaluations demonstrate the superiority of our method on reconstruction from point cloud, generation, and interpolation.
\end{abstract}

\begin{IEEEkeywords}
Structure Parameterize and Template; Structure Analysis; 3D Shape Reconstruction and Generation; 3D Shape Interpolation.
\end{IEEEkeywords}

\section{Introduction}

\IEEEPARstart{I}{n} recent years, the need for 3D models has grown with the improvement of various 3D applications. Therefore, reconstructing models from point clouds and generating new models have become crucial problems. In addition, some applications require that models contain structural semantic information and are easy to manipulate. A feasible approach is to represent objects structurally \cite{StructuralProcessing} for part-aware reconstruction and generation, which is also a hot research topic recently. The key to this approach is abstracting the shape of the object as cuboids (also called parts or boxes).

Different from 3D shape generation works that directly generate the surfaces and meshes of objects,
several recent works utilize hierarchical structural \cite{GRASS, StructureNet} or program-based \cite{ShapeAssembly} representations for shape generation and reconstruction in the form of cuboids and achieve great performance.
However, these methods either require hierarchical annotations or lack constraints on component relationships, leading to complicated neural networks for application or unreasonable generated shapes. 
Besides, these methods do not pay sufficient attention to the details inside cuboids.
They employ a network to learn the details and represent them with voxels or point clouds.
The storage and computational requirements of utilizing voxels are considerable. 
Utilizing point clouds also makes it hard to obtain meshes for downstream applications.

We observe that objects of the same category often share similar structures, a fact that should be exploited for taking advantage of structural information to represent objects. For example, a chair typically consists of one backrest, one seat, and four legs, and the legs of various chairs often share similar shapes, each being less complex than the entire chair. This motivates us to study how shared structural information of a category could benefit the representation of objects and their details of parts.

Inspired by this, we introduce a method designed to parameterize structures of shapes with differentiable templates for different categories and generate diverse parameters for 3D shape generation.
Different from previous works where each model has its own structure defined through various approaches, we design a differentiable template of a shared structure for each category and parameterize the shape based on the template, leading to fixed-length parameters.
In detail, the differentiable template is defined according to the configuration that records the constrictions and relationships of cuboids of a category. 
The template is implemented using a computation graph to delineate the process of differentiable calculations from specific parameters to the shapes that are represented as  combinations of cuboids.
To further represent the details inside each cuboid, we employ the boundaries of three-view drawings, which can be directly obtained from point clouds and are easy to learn and generate for neural networks.
The objects can be easily recovered by calculating the SDF according to parameters and details.
Benefiting from our fixed-length parameters and three-view details, our networks for reconstruction and generation, where only MLPs are employed, are simple and effective to learn the latent space. 
The parameters can be optimized without supervision using the differentiable template, which benefits us in building our dataset. The dataset comprises point clouds and corresponding manually annotated parameters for training a neural network to predict the parameters from point clouds and generate new shapes.
With well-structured latent space, our method can interpolate shapes between two objects.
Figure \ref{fig:teaser} shows several reconstruction, generation, and interpolation results of our method.

Our method has been rigorously validated, showing its superior ability to represent, reconstruct and generate diverse shapes.
The smooth interpolation results also demonstrate the rationality of our representation approach which can be learned well by networks.  
We will make our code and dataset publicly available.

Our main contributions are as follows.
\begin{itemize}
    \item We propose a novel method to parameterize structure with differentiable template for 3D shape generation. Our method defines a generic template for a category through a computation graph and describes the shapes of objects through specific parameters.
 
    \item We employ the boundaries of three-view drawings for further description of the inside details.  

    \item We train networks based on our method to reconstruct shapes from point clouds, generate diverse new shapes with complicated details and interpolate them.

\end{itemize}
\section{Related Work}

\begin{table}[]
\caption{The comparison of different methods. ``Part BBox'' indicates the oriented bounding box of each part, where parts are segmented according to the mesh parts from ShapeNet \cite{chang2015shapenet} or hierarchical part labels from PartNet \cite{mo2019partnet}.}
\label{tab:compare_relate}
\footnotesize
\centering
\setlength{\tabcolsep}{3pt}
\begin{tabular}{r|cccc}
\toprule
 & GRASS & StructureNet & ShapeAssembly & Ours \\ \midrule
\makecell[r]{Data \\ Requirement} & \makecell{Part BBox \\ (ShapeNet)} & \makecell{Hierarchical \\ Annotatio \\ (PartNet) } & \makecell{Part BBox \\ (PartNet)} & \makecell{No \\ Requirement} \\
\midrule
\makecell[r]{Network \\ Architecture} & \makecell{Recurrent \\ Network} & \makecell{Graph \\ Network} & \makecell{Recurrent \\ Network} & MLP \\ 
\midrule
\makecell[r]{Representation} & Cuboids & Graph & Programs & Parameters \\ 
\midrule
\makecell[r]{Detail } & Voxel & \makecell{Point \\ Cloud} & \makecell{Point \\ Cloud} & \makecell{Three-view \\ Boundaries} \\ 
\midrule
\makecell[r]{Mesh Result} & \Checkmark  & \XSolidBrush & \XSolidBrush & \Checkmark \\ 
\midrule
\makecell[r]{Optimization} & \XSolidBrush  & \XSolidBrush & \Checkmark & \Checkmark \\
\midrule
\makecell[r]{Reconstruction} & \XSolidBrush & \Checkmark & \Checkmark & \Checkmark \\ 
\midrule
\makecell[r]{Semantic} & \XSolidBrush & \Checkmark & \XSolidBrush & \Checkmark \\ 
\bottomrule
\end{tabular}
\end{table}

\subsection{Shape Parametrization and Representation}
CAD parametric modeling, widely used in industrial design, employs specific parameters of primitives and operations to represent objects accurately.
Various approaches \cite{li2023ConeConstruction, li2022integerconstrained, shen2022crossfields} utilize diverse representation methods including cone singularity construction and prescribed holonomy signatures to parameterize object surfaces.
Methods like SCAPE \cite{SCAPE_human_body_2005}, SMPL \cite{SMPL_human_body_2015}, and SMPL-X \cite{SMPL_X_human_body_2019}  introduce specific parameters to control the human body poses and surfaces, for human bodies reconstruction.
These methods are designed to accurately represent the geometric surfaces of objects or human bodies, while our method parameterizes the structure of objects, providing an abstract representation.

Many methods utilize different 3D primitives to represent objects \cite{SurveyPrimitive}, such as cuboids\cite{Kim13LearningPartBaseTemp}, cones\cite{GlobFit}, spheres\cite{TGB13} and so on. 
Further methods \cite{MetaRepre, ShapeTemplates} introduce part semantic information on primitive cuboids for structure analysis.
Procedural modeling works \cite{BayesianGrammar, PMBuildings, ProceduralModeling} represent objects utilizing programs in high-level approaches.
These works are designed for fitting the surfaces without considering the semantic information of each primitive. 
They are also hard to be applied in neural networks for applications such as reconstruction and generation.

\begin{figure*}[t]
    \centering
    \includegraphics[width=0.9\linewidth]{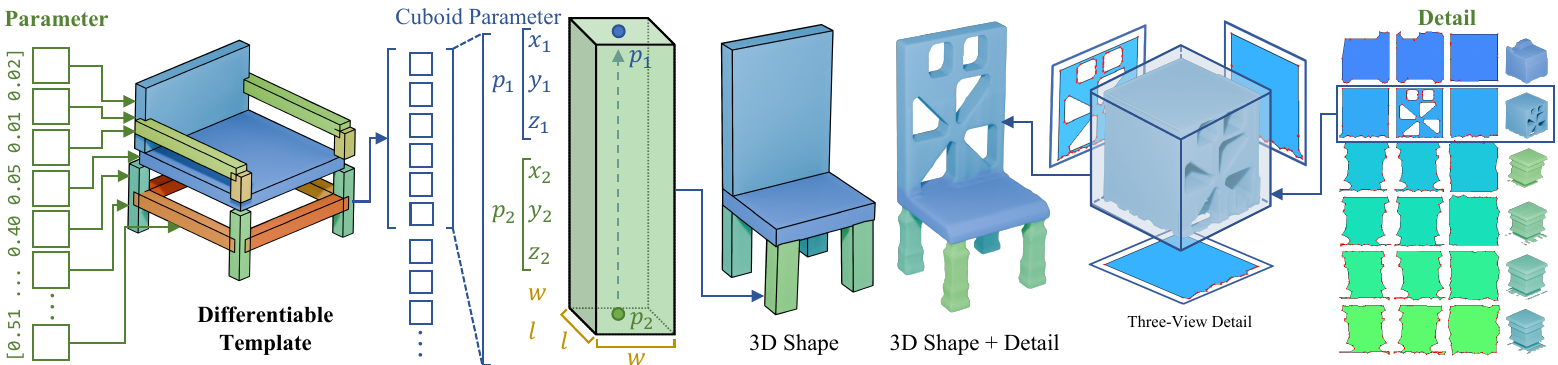}
    \caption{We utilize parameters to control the position of cuboids through a differentiable template. Each cuboid is defined as a ``stick'' with two control points and sizes. The object  is represented by the combination of cuboids. We employ three-view boundaries (black contours) to represent the details inside each cuboid. Here, we only store the red vertexes.}
    \label{fig:arg_model}
\end{figure*}

\subsection{Shape Generation with Cuboids}
Our work is mainly related to works based on deep learning for generating 3D shapes represented in cuboids. 
GRASS \cite{GRASS} introduces a generative recursive autoencoder for shapes. The autoencoder extracts symmetry hierarchies from unlabeled cuboids unsupervisedly. This method generates new models based on the autoencoder and a generative network for volumetric part geometries.
Mo et al. propose StructureNet \cite{StructureNet}, a hierarchical graph network for learning a unified latent space for shapes. They employ the $n$-array graph to represent the shapes and utilize graphic networks to achieve the generation of new shapes and point clouds. Different from GRASS, the generated results of StructureNet contain semantic information about each cuboid.
ShapeAssembly \cite{ShapeAssembly} executes programs that constrict the relationship with the grammar, to produce cuboids for representing shapes. A recurrent network is employed to encode programs into latent space and generate new programs for new shapes. 
For different purposes, ShapeMOD \cite{ShapeMOD} and ShapeCoder \cite{ShapeCoder} are proposed to discover the macro operations for simplifying the program and reducing the difficulty of editing. 
Though these methods perform well for shape generation with cuboids, they still have some limitations. 
These methods rely on hierarchical part annotations that are time-consuming to obtain. They usually employ recurrent and graph networks as basic modules due to irregular representation approaches. Besides, these methods pay less attention to the details inside cuboids. 
Different from them, our method employs networks based on MLPs without requiring part annotations. Besides generating, our method can also optimize or reconstruct the shapes from point clouds. Combining with three-view boundaries of details, mesh results can be efficiently obtained through our method.
Table \ref{tab:compare_relate} summarizes the differences between our method and previous methods.

\subsection{Shape Generation and Reconstruction with Mesh}
Shape generation methods \cite{mittal2022autosdf, cheng2023sdfusion} generate holistic meshes with images or texts as conditions for objects.
Part-aware shape generation methods focus on generating the surface for each part, based on different approaches that generates shapes with cuboids.
These methods take less consideration of generation shapes with cuboids.
SDM-Net \cite{SDM-NET} and DSG-Net \cite{DSG-Net} are based on GRASS and StrcutureNet,  separately. They employ part mesh modules to encode and decode the geometry of parts. 
SDM-Net combines all latent codes of parts and utilizes VAE to generate new objects. 
DSG-Net employs a weight-shared conditional-VAE and takes the cuboids features as conditions to generate new objects.
Different from them, SALAD \cite{SALAD} represents shapes as Gaussians \cite{pointgmm2020, hertz2022spaghetti} and generates surfaces without decomposing each part through a diffusion approach.


Surface reconstruction methods including traditional approaches \cite{bernardini1999BPA, kazhdan2006poisson, kazhdan2013possionscreened}, optimization-based methods \cite{hanocka2020point2mesh, ma2022PCP, lin2022PGR} and deep-learning-based methods \cite{huang2022NGS, boulch2022poco, Ren_2023_ICCV}, focus on reconstructing the whole surface of the object from its point cloud. 
Our method focuses on reconstructing the abstracted cuboids from a point cloud and refining details inside each cuboid.
The targets of shape generation and reconstruction with mesh in this section are different from the goals of our method.

\section{Method}

In this section, we first introduce the representation approach of cuboids and details in Section \ref{sec:rep}. Then we introduce the parameterization of shapes in Section \ref{sec:para_structure}, including the differentiable template and the definition of the template. Finally, we introduce the approaches utilized in reconstruction and generation.

\subsection{Representation of Cuboids and Details} \label{sec:rep}

We utilize a group of cuboids to represent the shape of an object as shown in the Figure \ref{fig:arg_model}. 
To define a cuboid, previous works utilize the transform matrix which is not intuitive for users to modify the position of the cuboid by adjusting its values.
Having too many degrees of freedom also makes it difficult to optimize.
In this paper, we employ an intuitive definition of cuboids by representing transform matrixes with cuboid parameters.
We define a cuboid as a ``stick'' with 8 cuboid parameters: $(x_1, y_1, z_1)$, $(x_2, y_2, z_2)$ for control point $p_1$ and $p_2$ that constrains its direction and height, and $w$, $l$ for its size. 
We set the rotation angle of cuboids along $p_1$ and $p_2$ to zero in practice.
To get the cuboid $B$, we transfer the unit cube with transform matrix $M_B$ which is easy to obtain from cuboid parameters through a differentiable calculation.
Detailed calculations can be found in the supplementary material.
Benefiting from this definition, we can intuitively understand how to move $p_1$ and $p_2$ or adjust $w$, $l$ for modifying a cuboid. 

A group of cuboids is not enough to represent the shape of an object.  
More details are required. 
Different from previous methods that employ the latent features learned by networks to describe the details inside cuboids, we directly utilize the pattern boundaries of three-view drawing to represent the detailed shape inside each cuboid as shown in Figure \ref{fig:arg_model}. 
These boundaries are non-convex polygons that may contain holes, and we only record their vertices to describe the detail inside a cuboid. 
Utilizing three-view details can save a lot of storage space. 
According to our statistics, storing meshes takes ten times more space than storing point clouds, while storing point clouds takes ten times more space than our method.
Meanwhile, three-view details are easy for neural networks to learn and generate,
which can be achieved by easily employing MLPs.


To recover the whole mesh of an object given the cuboids and corresponding details, we employ Algorithm \ref{alg:alg1} to get the SDF and apply the marching cubes algorithm \cite{lorensen1998marchingcube} on it. 
Here, $\bm{D}[i]_x, \bm{D}[i]_y, \bm{D}[i]_z$ indicates the boundaries of three-view details of the $i$-th cuboids, and $M_{B[i]}^{-1}p$ denotes the inverse transform matrix of the $i$-th cuboid.
Given a query point, the algorithm first finds the cuboid that the point belongs to and then projects the point on three-view drawings to calculate the distance to the boundaries. 
The minimum distance is the SDF value of the query point.

\begin{algorithm}[t]
\caption{}\label{alg:alg1}
\small
\begin{algorithmic}[1]
\Require Cuboids $\bm{B}$, Three-view details $\bm{D}$, Resolution $R$, \textbf{in}($p$, $d$): whether point $p$ is inside polygon of detail $d$, \textbf{dis}($p$, $d$): the distance from point $p$ to polygon of detail $d$. 
\Ensure SDF $\bm{V}$
\State $\bm{V}$ $\gets$ the volumes of SDF with resolution $R$
\For{volume $v$ in $\bm{V}$}
\State $p$ $\gets$ $v$.\textit{coordinate}
\State $i$ $\gets$ the index such that $p$ is inside $\bm{B}[i]$
\If{$i$ is not exist}
\State $v$.\textit{value} $\gets$ 1
\Else 
\State $p$ $\gets$ $M_{B[i]}^{-1}p$
\State $p_x, p_y, p_z$ $\gets$ projection $p$ to yz, xz, xy plane
\State \textit{flag} $\gets$ \textbf{in}($p_x$, $\bm{D}[i]_x$) \textbf{and} \textbf{in}($p_y$, $\bm{D}[i]_y$) \textbf{and} \textbf{in}($p_z$, $\bm{D}[i]_z$) 
\State \textit{dis} $\gets$ \textbf{min}(\textbf{dis}($p_x$, $\bm{D}[i]_x$), \textbf{dis}($p_y$, $\bm{D}[i]_y$), \textbf{dis}($p_z$, $\bm{D}[i]_z$))
\State $v$.\textit{value} $\gets$ $-$\textit{dis} \textbf{if} \textit{flag} \textbf{else} \textit{dis}
\EndIf
\EndFor
\end{algorithmic}
\end{algorithm}

\subsection{Parameterization of Shape} \label{sec:para_structure}
\subsubsection{Differentiable Computation Graph of Template}\label{sec:dcg}

In a category, objects usually have similar structures with certain characteristics.   
Take a chair as an example, its legs always connect with the seat and are always left-right symmetry with each other.
Because of the existence of these relations, not all cuboids are completely free. Some of the cuboid parameters rely on others.
As shown in Figure \ref{fig:arg_model}, 
we employ a differentiable template to calculate the cuboid parameters from fewer parameters.
In practice, the template is implemented using a differentiable computation graph.
The computation of the $i$-th cuboid parameter $b_i$ is defined as:
\begin{equation}\label{equ:main}\small
b_i = 
\underset{\textcircled{1}}{\underline{c_i}} + 
\underset{\textcircled{2}}{\underline{r_i {\color{red} a_{i1}}}} + 
\underset{\textcircled{3}}{\underline{s_{i1}\bm{K}_{j_1k_1}\bm{e_1}}} + 
\underset{\textcircled{4}}{\underline{s_{i2}({\color{red} a_{i2}}\bm{K}_{j_2k_2}+(1-{\color{red} a_{i2}})\bm{K}_{j_3k_3})\bm{e}_2}}
.
\end{equation}
Here, $a_{i1}, a_{i2}\in [0, 1]$, $c_i, r_i \in \mathbb{R}$, $s_{i1}, s_{i2} \in \{-1, 0, 1\}$, $j_1, j_2, j_3 \leq i$. 
$\bm{K}_{jk}$ indicates the $k$-th key point of the $j$-th cuboid. We define 26 key points for each cuboid, including the center points of 6 faces, the 8 vertices, and the midpoints of 12 edges. The key points can be derived once the transform matrix of the cuboid is obtained.
Detailed calculations can be found in the supplementary material.
$\bm{e_1}, \bm{e_2}$ represents one of the basic vectors of the x, y, z axes, that is $[1, 0, 0]^T, [0, 1, 0]^T, [0, 0, 1]^T$. 
Different terms indicate different relationships: \textcircled{1} indicates a fixed offset; \textcircled{2} indicates an offset controlled by a parameter; \textcircled{3} indicates $b_i$ is related to a key point of $j_1$-th cuboid; \textcircled{4} indicates $b_i$ is related to a line connecting two key points. 
Only $a_{i1}, a_{i2}$ are parameters, and other variables are specified according to the configuration of the template. 
The number of parameters for one category is fixed.
Equation \ref{equ:main} is differentiable. Therefore, according to the chain rule, the computation graph from parameters to cuboid parameters is also differentiable. 
In practice, most cuboid parameters have no term \textcircled{2} or \textcircled{4}. 
The number of parameters is fewer than cuboid parameters.
Different objects of a category are represented in different values of the fixed-length parameters given from a template and corresponding details (boundaries of three-view drawings).
If the cuboid's volume of a shape is smaller than a specified threshold, we will remove it.

\subsubsection{Configuration of Template}

\begin{figure}[t]
    \centering

    \includegraphics[width=\linewidth]{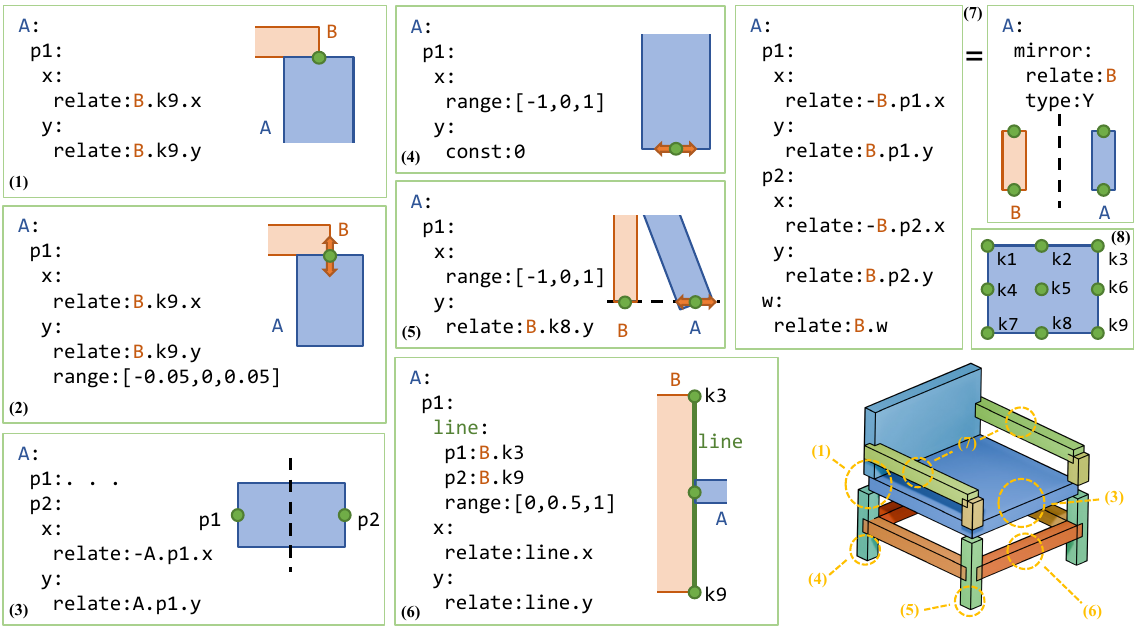}
    \caption{Several relationships are utilized in differentiable templates including joint (1)(2), restriction (4)(5), line (6), and symmetry (3)(7), in the 2D version. An example (Chair-4) is shown at the bottom right to illustrate the utilization of these relationships.}
    \label{fig:design}
\end{figure}

Defining cuboids as sticks makes it easy to define the relationship between cuboids, such as joint connection and symmetry.
The control points advance in defining the attachment relationship of cuboids.
Figure \ref{fig:design} illustrates the basic relationships used by template configurations in the 2D version and a template example.  
(1), (2) describe a connection of two cuboids, such as the joint of the leg and the seat. (2) offers a slight offset for the joint for more precise representation.
(3) represents the symmetry of one cuboid, for example, the left-right symmetry of the seat.
(4), (5) restrict control point to a line or other cuboid, which can be utilized to require that all legs of a chair have the same height.
(6) illustrates the ``line'' mentioned in the term \textcircled{4} of Equation \ref{equ:main}. Restricting the control point of the horizontal bar to move on the leg can be achieved through this relationship.
(7) shows the symmetry between two cuboids, such as the arms.
How to setting the variables in Equation \ref{equ:main} can be found in the supplementary material.
We also provide a tool for users to visually, interactively, and intuitively design the template for a category. 
The template can also be designed from the hierarchical annotation \cite{StructureNet} of objects automatically. 
More examples of the configurations of templates, more details of our tool and automatic template design can be found in the supplementary material. 

\subsection{Reconstruction and Generation}
\subsubsection{Reconstruction}\label{sec:recon}
Shape reconstruction is an important task that can not only produce a 3D model from an easily obtained point cloud but also measure the representational ability of a method.  
Given the template of a category, there are two approaches to reconstruct the shape of an object from its point cloud $\mathcal{P}$, concretely, that is predicting the parameters according to $\mathcal{P}$.
One is an optimization-based approach. 
We sample the cuboids into point cloud $\mathcal{P}_s$ according to their transform matrices, which is a differentiable process.
Detailed calculations can be found in the supplementary material.
And from Sections \ref{sec:rep} and \ref{sec:dcg} we know that the computation from parameters to cuboid is also differentiable.
Thus, the whole process is differentiable and gradient descent can be adopted to optimize parameters for minimizing the Chamfer Distance \cite{fan2017point_cd} between $\mathcal{P}$ and $\mathcal{P}_s$. 
Another one is a data-driven approach utilizing an encoder-decoder neural network. 
Different from previous works that are mainly RNN-based networks with complicated architecture, 
we employ PointNet++ \cite{qi2017pointnet++} as an encoder to extract features from point clouds and MLPs as a decoder to simply predict parameters from extracted features, 
since the length of parameters is fixed for all objects in a category.
After obtaining the parameters of an object, we split out the points inside each cuboid to reconstruct the details inside. 
For each cuboid, we project its inside points to the three-view drawing and apply AlphaShape \cite{alphashape} to recover the boundaries of the three-view drawing from 2D points.
Finally, we acquire the parameters and details for an object from its point cloud. 
More details of our reconstruction approach can be found in the supplementary material.

\subsubsection{Generation}
We employ VAE \cite{VAE_Kingma2013AutoEncodingVB} to learn the latent space of parameters for a category and generate new parameters.
The networks for details utilize the same architecture, except that the inputs are images.
All encoders and decoders are MLPs. 
We redraw the boundaries to binary images where the inside areas are filled with ``1''.
The input and output of VAE are images. After generating new images, we extract their boundaries as newly generated details.
To keep the symmetry of the generated model, we reflect the symmetrical part that is already generated according to the template. 
For example, we first generate the left arm of a chair. Then, we reflect the left arm and place it in the position of the right arm rather than generate a new one.


\section{Results and Evaluation}

\subsection{Data}


We select nearly 4,000 models from the ShapeNet dataset \cite{chang2015shapenet}, a collection of 3D CAD models, and classify the models into 20 categories.
We also design the templates for 20 categories and annotate the parameters for all models.
With our visualization tools, designing a template for a category takes about 10-20 minutes.
The optimization-based method mentioned in Section \ref{sec:recon} plays a crucial role in annotation. 
We only need to fine-tune the optimized initial parameters to obtain the ground truth parameters, reducing nearly 50\% - 60\% annotation time from 3-6 minutes to 1-2 minutes.
We sample models to point clouds together with parameters as the training data for reconstructing parameters from point clouds.  
Based on the annotated parameters of each model, we can easily obtain the details. 
The parameters and details are used to train the VAE to generate new shapes.
We randomly select 200 models for testing.

\subsection{Reconstruction}

\begin{figure*}[t]
    \centering
    \includegraphics[width=\linewidth]{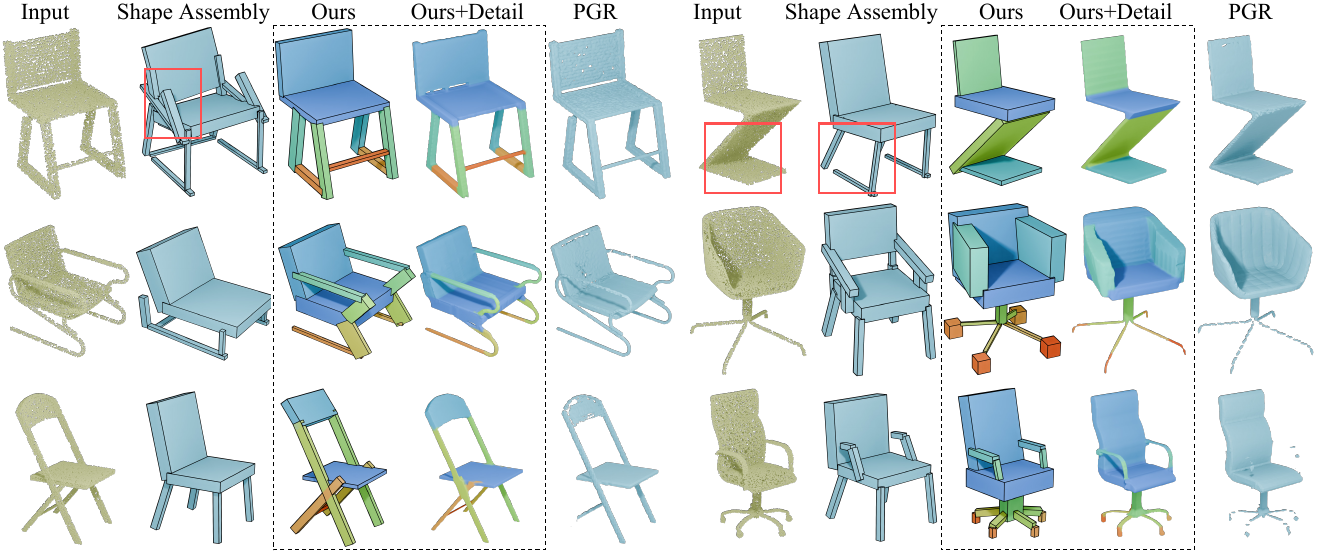}
    \caption{Reconstruction results from point clouds of different methods on Chair category. Here, we display the reconstruction results of a surface reconstruction method (PGR) for reference to evaluate the quality of our reconstructed shapes with details.}
    \label{fig:recon}
\end{figure*}

\subsubsection{Data-driven approach}
To show the representation ability of the proposed method, we compare our method with ShapeAssembly \cite{ShapeAssembly} on reconstructing shape from point cloud. 
The comparison of the Chair category is shown in Table \ref{tab:recon_com}. 
We use the meshes generated by Algorithm \ref{alg:alg1} from parameters and details as the results of ``Ours+Detail'', to verify the necessity of detail representation.
We employ three metrics to measure the distance from the reconstructed shape and target point cloud $\mathcal{P}_t$, including surface distance, solid distance, and symmetry distance.
The surface distance is the Chamfer Distance between the point clouds sampled from the surfaces ground truth mesh and the reconstructed mesh. 
The solid distance is the Chamfer Distance between the point clouds sampled from the inside area of the ground truth mesh and the reconstructed mesh. 
Reflecting the overlap of the inner regions of the two shapes, this distance provides a better measure of the accuracy of the reconstruction of the shape.
Symmetry distance measures the symmetry of $\mathcal{P}_s$ when $\mathcal{P}_t$ is symmetric about yz, xz, xy planes:
$$
\frac{1}{3}\hspace{1mm}\sum_{\mathclap{p\in \{yz, xz, xy\}}}\hspace{1mm} \sqrt{\left(\log{\bm{CD}\left(\mathcal{P}_s, \mathcal{P}_s^{(p)}\right)}-\log{\bm{CD}\left(\mathcal{P}_t, \mathcal{P}_t^{(p)}\right)}\right)^2},
$$
where $\mathcal{P}^{(p)}$ indicates the symmetric point cloud of $\mathcal{P}$ about plane $p$.
For shape reconstruction with cuboids, we take the meshes of cuboids as reconstruction results.
The results show that, even though our method utilizes one template for models of a whole category, our method still performs better than ShapeAssembly, showing the superiority of our representation approach. 
This also benefits from the fixed-length parameters and simple network that is easy to train. 
The qualitative comparison is shown in Figure \ref{fig:recon}. 
Our method can accurately reconstruct complicated shapes from point clouds, such as the armrests and the uncommon ``Z'' leg structures. 
With the detail of each cuboid, our method can further represent more detailed shapes inside each cuboid. 
The curved armrests and legs of chairs are represented precisely, even for the complicated patterns of the backrest shown in Figure \ref{fig:teaser}.

To show the reconstruction ability of our method, we also conduct a comparison with the surface reconstruction method PGR\cite{lin2022PGR} for reference.
Even though our method is not designed for surface reconstruction, our method still performs better than PGR in some categories, such as Swivel Chair-5 or Folding Chair on some metrics as shown in Table \ref{tab:recon_com}, further indicating the superiority of our method.
The qualitative results also verify that combining structural cuboids and three-view boundaries can represent detailed shapes accurately since the shape inside each cuboid is simple enough for three-view.

\begin{table}[]
\caption{Comparison of different methods on reconstruction from point cloud. The best results of all methods are represented in bold. The best results of shape reconstruction methods are indicated with  \squarefirst backgrounds. The results of surface loss and inside loss are multiplied by 100, and the results of symmetric loss are multiplied by 10.}
\label{tab:recon_com}
\small 
\centering
\setlength{\tabcolsep}{4pt}
\begin{tabular}{@{}r|l|ccc@{}}
\toprule
 & Method & Surface $\downarrow$ & Solid $\downarrow$ & Symmetric $\downarrow$ \\ \midrule
 & ShapeAssembly & 1.964 & 3.009 & 5.013 \\
 & Ours & 0.551 & 1.303 & 2.797 \\
 & Ours+Detail  & \colorfirst0.171 & \colorfirst1.067 & \colorfirst2.171 \\
\multirow{-4}{*}{Chair-4} & PGR & { \textbf{0.135}} & { \textbf{0.862}} & { \textbf{0.943}} \\ \midrule
 & ShapeAssembly & 8.016 & 6.766 & 6.720 \\
 & Ours & 1.660 & 2.033 & 3.750 \\
 & Ours+Detail & \colorfirst0.307 & \colorfirst\textbf{1.828} & \colorfirst2.358 \\
\multirow{-4}{*}{\begin{tabular}[c]{@{}r@{}}Swivel\\ Chair-5\end{tabular}} & PGR & { \textbf{0.048}} & { 1.906} & { \textbf{1.536}} \\ \midrule
 & ShapeAssembly & 5.286 & 4.753 & 9.394 \\
 & Ours & 2.198 & 1.862 & 5.401 \\
 & Ours+Detail & \colorfirst0.765 & \colorfirst1.727 & \colorfirst1.959 \\
\multirow{-4}{*}{\begin{tabular}[c]{@{}r@{}}Swivel\\ Chair-4\end{tabular}} & PGR & { \textbf{0.065}} & { \textbf{1.613}} & { \textbf{0.886}} \\ \midrule
 & ShapeAssembly & 3.850 & 4.841 & 5.114 \\
 & Ours & 1.301 & 1.597 & 4.134 \\
 & Ours+Detail & \colorfirst0.239 & \colorfirst\textbf{1.213} & \colorfirst1.470 \\
\multirow{-4}{*}{\begin{tabular}[c]{@{}r@{}}Cantilever\\ Chair\end{tabular}} & PGR & { \textbf{0.085}} & { 2.408} & { \textbf{1.210}} \\ \midrule
 & ShapeAssembly & 5.599 & 8.711 & 7.102 \\
 & Ours & 1.129 & \colorfirst\textbf{4.965} & 3.065 \\
 & Ours+Detail & \colorfirst\textbf{0.131} & 5.584 & \colorfirst\textbf{1.622} \\
\multirow{-4}{*}{\begin{tabular}[c]{@{}r@{}}Folding\\ Chair\end{tabular}} & PGR & { 0.146} & { 5.272} & { 2.996} \\ \midrule
 & ShapeAssembly & 4.586 & 4.977 & 4.750 \\
 & Ours & 1.326 & 1.228 & 3.064 \\
 & Ours+Detail & \colorfirst0.251 & \colorfirst\textbf{0.766} & \colorfirst\textbf{1.223} \\
\multirow{-4}{*}{\begin{tabular}[c]{@{}r@{}}Sculptural\\ Chair\end{tabular}} & PGR & { \textbf{0.195}} & { 0.788} & { 1.579} \\ \midrule
 & ShapeAssembly & 4.883 & 5.509 & 6.349 \\
 & Ours & 1.361 & 2.165 & 3.702 \\
 & Ours+Detail & \colorfirst0.311 & \colorfirst\textbf{2.031} & \colorfirst1.801 \\
\multirow{-4}{*}{AVG.} & PGR & { \textbf{0.112}} & { 2.141} & { \textbf{1.525}} \\ \bottomrule
\end{tabular}
\end{table}


\begin{figure}[t]
    \centering
    \includegraphics[width=0.8\linewidth]{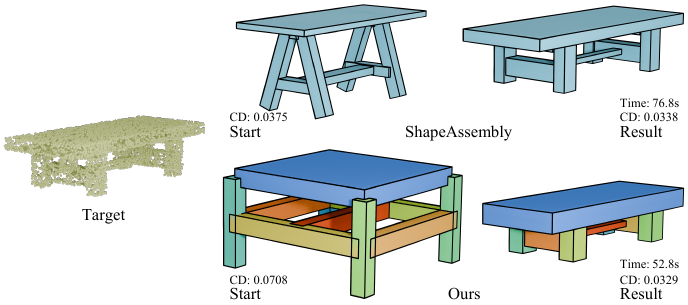}
    \caption{An optimization example on Table-4 category of our method and ShapeAssembly. Each shape is optimized 1000 iterations.}
    \label{fig:opt}
\end{figure}

\begin{figure*}[t]
    \centering
    \includegraphics[width=\linewidth]{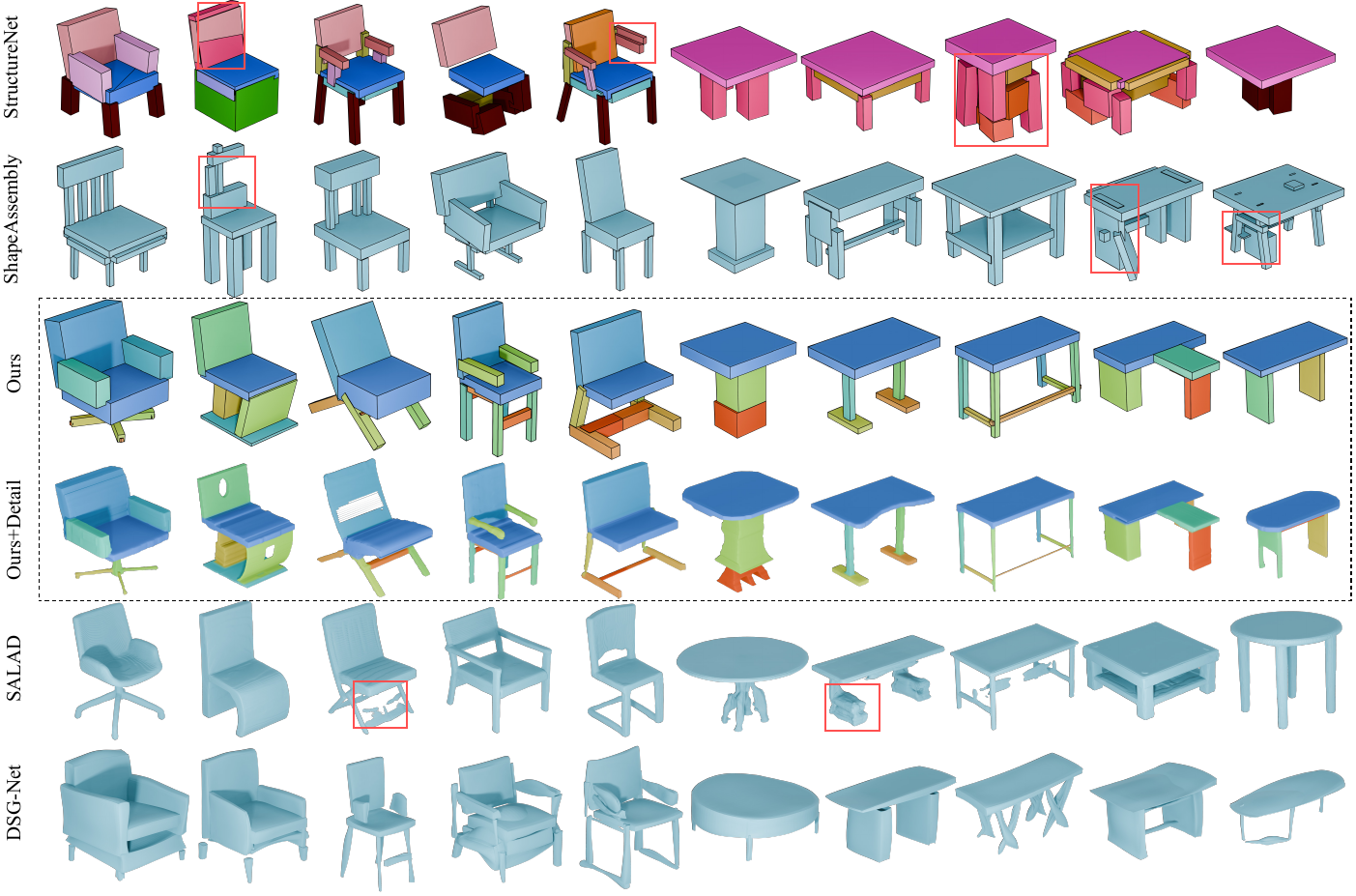}
    \caption{The generated shapes of our method, StructreNet, ShapeAssembly, SALAD, and DSG-Net on Chair and Table categories.}
    \label{fig:gen}
\end{figure*}

\subsubsection{Optimization-based approach}
We also conduct an experiment to verify the efficiency of the proposed optimization-based approach, by optimizing a shape from the target point cloud unsupervisedly. 
We compare our method with ShapeAssembly in Table-4 category.
The average Chamfer Distances between the optimized shapes and target point clouds are 0.127 for ShapeAssembly and 0.017 for our method.
Though ShapeAssembly can optimize a shape based on a given program of another similar object, its performance is worse than ours since it represents an object with one program while our proposed method represents a shared common structure for a category.
As shown in Figure \ref{fig:opt}, our method takes less time with more accurate shapes.



\subsection{Parameter Number}

\begin{table}[]
\caption{Comparison of different methods on Para-Part Ratio. The first three minor values in each column are demarcated by backgrounds, which are colored \squarefirst, \squaresecond, and \squarethird.}
\label{tab:para-part-ratio}
\small
\centering
\setlength{\tabcolsep}{1.3mm}{
\begin{tabular}{@{}r|ccc|ccc@{}}
\toprule
\multicolumn{1}{l|}{} & \multicolumn{3}{c|}{Chair} & \multicolumn{3}{c}{Table} \\ \midrule
Method & Para & Part & Ratio $\downarrow$ & Para & Part & Ratio $\downarrow$ \\ \midrule
StructureNet & 119.29 & 13.25 & 9.00 & 101.66 & 11.30 & 9.00 \\
ShapeAssembly & 113.50 & \colorsecond11.60 & 9.78 & 96.60 & \colorthird9.44 & 10.23 \\
ShapeMod & 73.10 & \colorsecond11.60 & 6.30 & 59.40 & \colorthird9.44 & 6.29 \\
ShapeCoder & \colorfirst27.00 & \colorfirst10.00 & \colorsecond2.70 & \colorsecond18.00 & \colorsecond8.00 & \colorsecond2.25 \\ \midrule
Ours & \colorthird39.80 & 13.75 & \colorthird2.90 & \colorthird22.55 & \colorfirst6.14 & \colorthird3.67 \\
Ours+ & \colorsecond27.54 & 13.28 & \colorfirst2.07 & \colorfirst17.89 & 9.65 & \colorfirst1.85 \\ \bottomrule
\end{tabular}
}
\end{table}

We employ the ratio of the average parameter number and average cuboid number of models, which is referred as para-part ratio, to show how many parameters are required to represent a cuboid of a model.
The comparison also shows the representational efficiency of different methods.
We compare the para-part ratio of the proposed method against StructureNet, ShapeAssembly, ShapeMOD, and ShapeCoder on Chair and Table, as shown in Table \ref{tab:para-part-ratio}. Note that ShapeMOD and ShapeCoder are methods focusing on discovering abstract patterns to decrease the number of parameters.
The results show that the number of cuboids used by our method to represent shapes is almost the same as other methods, while the number of parameters of our method is less. 
Our method has lower para-part ratios than ShapeAssembly while its reconstruction performance is better than ShapeAssembly, showing the efficiency of our representation approach.
Even though our method uses fewer parameters, its reconstruction performance is still better than compared methods.
``Ours+'' indicates the automatically designed template from hierarchical annotation, where each model has its own template. 
In this situation, our method even performs better than ShapeMOD and ShapeCoder.


\begin{figure*}[t]
    \centering
    \includegraphics[width=\linewidth]{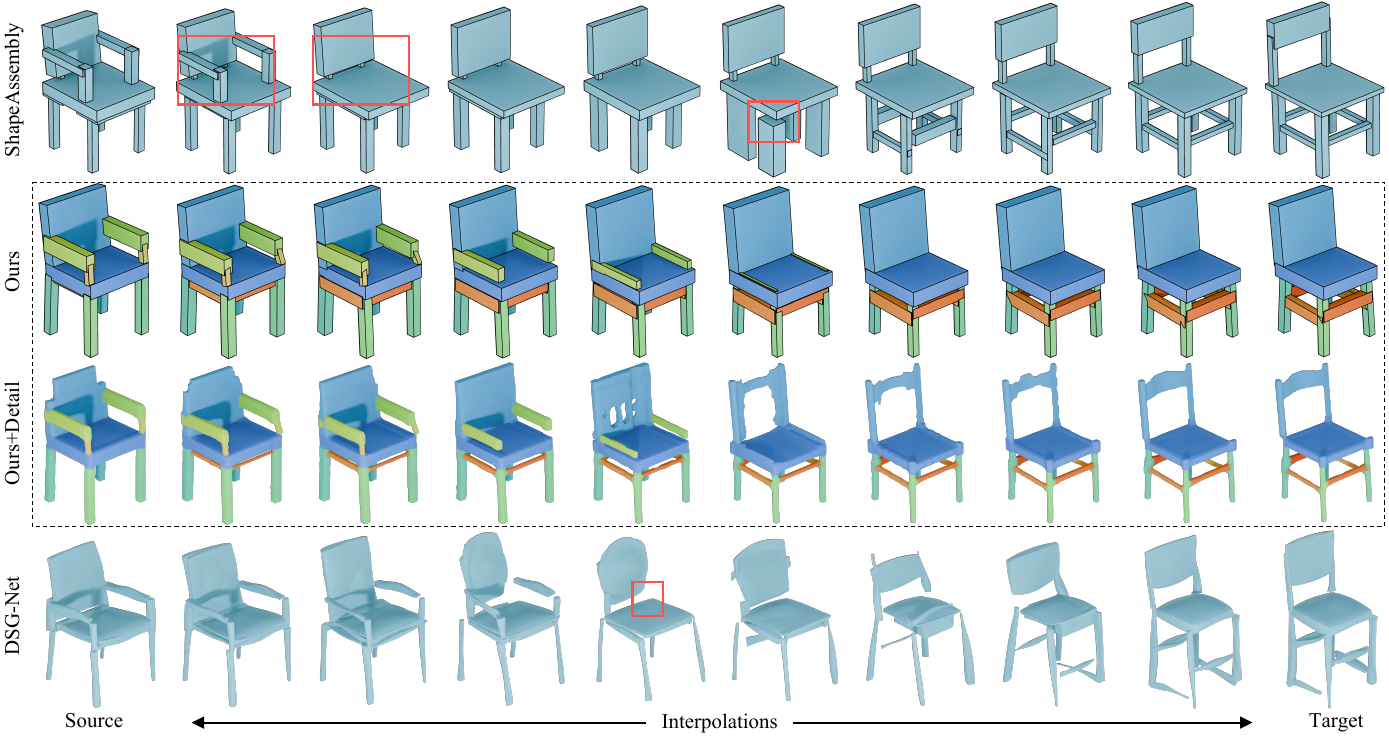}
    \caption{The interpolation results of our method, ShapeAssembly, and DSG-Net. The results show the smooth interpolation processes on the shapes of our method.}
    \label{fig:inter}
\end{figure*}

\subsection{Generation}
\subsubsection{Shape Generation with Cuboids}

\begin{table}[]
\caption{Comparison of different methods on shape generation. The best results are highlighted in bold.}
\label{tab:gen}
\small
\centering
\setlength{\tabcolsep}{8pt}
\begin{tabular}{@{}r|l|ccc@{}}
\toprule
\multicolumn{1}{l|}{} & Method & Rooted $\uparrow$ & Stability $\uparrow$ & Fool $\uparrow$ \\ \midrule
\multirow{3}{*}{Chair} & StructureNet & 89.7 & 74.9 & 4.04 \\
 & ShapeAssembly & 94.5 & 84.7 & 25.6 \\
 & Ours & \textbf{96.1} & \textbf{93.7} & \textbf{53.6} \\ \midrule
\multirow{3}{*}{Table} & StructureNet & 94.4 & 76.8 & 3.94 \\
 & ShapeAssembly & 96.2 & 85.9 & 33.2 \\
 & Ours & \textbf{99.8} & \textbf{87.1} & \textbf{51.3} \\ \bottomrule
\end{tabular}
\end{table}

We qualitatively and quantitatively compare our shape generation method with StructureNet and ShapeAssembly.
We employ three metrics \cite{ShapeAssembly}, including rooted, stability, and fool, on randomly generated 1000 models of Chair and Table category.
The results are shown in Table \ref{tab:gen}.
The relationships of cuboids of our shapes are explicit according to the template, and the geometric positions of cuboids are restricted reasonably.
Thus our generated shapes are more stable with fewer fly parts and easy to confuse the discriminator.

Figure \ref{fig:gen} shows the quantitative comparison of different methods in Chair and Table category, including StructureNet and ShapeAssembly.
As the results indicate, the generated results of StructureNet and ShapeAssembly may have irrational and redundant parts, such as the backrests of chairs and the legs of tables.
On the contrary, with a rigorous template of a category, the generated parts have a certain number and specific relationship. Thus, our generated results are reasonable with concise and complicated shapes.


\begin{figure}[ht]
    \centering
    \includegraphics[width=\linewidth]{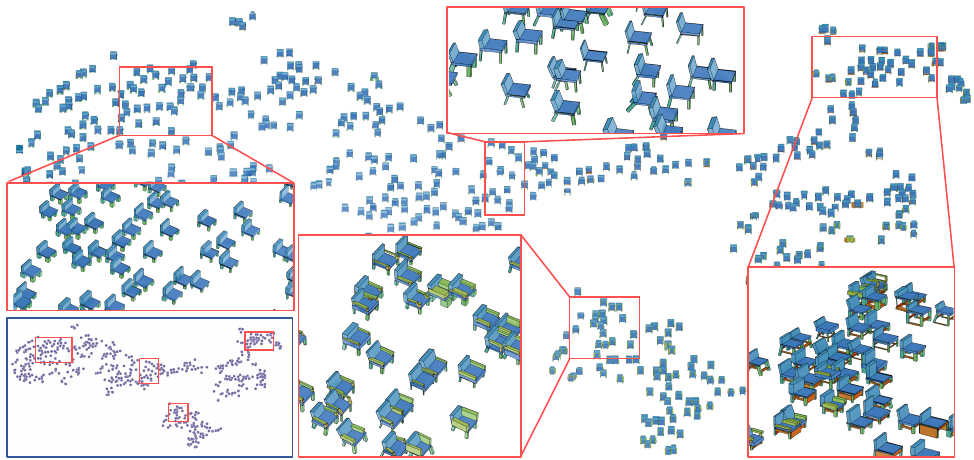}
    \caption{Visualization of shape latent codes generated by VAE on Chair-4 category, where similar shapes are clustered together.}
    \label{fig:latent}
\end{figure}

\subsubsection{Visualization of VAE's Latent Space}

In order to assess the efficacy of the parameters learned by the VAE, we utilize t-SNE to visualize the well-structured latent space. As is evident from Figure \ref{fig:latent}, the visualization clearly demonstrates the well-organized latent space.
The observed clustering of similar shapes, such as chairs with armrests or crossbars, is indicative of the proficiency of the VAE in learning the distinguishing features of different shapes. This points towards the successful acquisition of the inherent characteristics of diverse shapes by the VAE.
This also shows the rationality of shapes generated by our method.

\subsubsection{Generation with Details}
Figure \ref{fig:gen} shows our generated shapes with details.
We also display the generated models of DSG-Net \cite{DSG-Net} and SALAD \cite{SALAD} for reference, to demonstrate the quality of shapes with details generated by our method. 
These two methods are specially designed for part-aware shape generation with mesh. 
Note that our method is not designed for surface generation.
The results show that our method can generate diverse models based on the cuboids. 
For example, the complicated pattern on the backrest of the chair indicates the efficiency of the representation approach of details.
Even though the architecture of our generative network is simple, the network still performs well benefiting from our representation approach.

\subsection{Interpolation}
\subsubsection{Shape Interpolation with Cuboids}

\begin{table}[]
\caption{Comparison of different methods on interpolation. Here, ``Ours (code)'' indicates interpolating codes, and ``Ours (para)'' indicates interpolating parameters. }
\label{tab:inter}
\small
\centering
\setlength{\tabcolsep}{12pt}
\begin{tabular}{@{}r|l|cc@{}}
\toprule
\multicolumn{1}{l|}{} & Method & Chair & Table \\ \midrule
\multirow{4}{*}{\begin{tabular}[c]{@{}r@{}}Shape\\ (CD) $\downarrow$\end{tabular}} & StrcutureNet & 0.0384 & 0.0474 \\
 & ShapeAssembly & 0.0384 & 0.0389 \\
 & Ours (code) & 0.0212 & 0.0254 \\
 & Ours (para) & \textbf{0.0178} & \textbf{0.0202} \\ \midrule
\multirow{2}{*}{\begin{tabular}[c]{@{}r@{}}Parameter\\ (MSE) $\downarrow$\end{tabular}} & Ours (code) & 0.4944 & 0.4299 \\
 & Ours (para) & \textbf{0.4320} & \textbf{0.3644} \\ \bottomrule
\end{tabular}
\end{table}

There are two approaches to achieve interpolation. One is directly interpolating the parameters, another is interpolating the code in the latent space of VAE.
Figure \ref{fig:inter} shows the interpolated shapes of the second approach.
The smooth changes between the shapes indicate that the parameters of our method appropriately represent the shapes.
The similarity of the interpolation process on parameters and codes also verifies the rationality of our method, which is shown in the supplementary materials.
We also compare the shape interpolations of the proposed method with ShapeAssembly.
The interpolation process of our method is smoother than ShapeAssembly. 
Since ShapeAssembly has no strict restrictions on cuboids, the legs of the interpolated chair are not symmetric and the armrests suddenly disappear. 
We also qualitatively compare different interpolation approaches on Chair and Table category. The results are shown in Table \ref{tab:inter}, where we randomly generate 100 pairs of shapes and interpolate 100 steps for each pair.   
The smooth distance of parameters is the MSE between the adjacent interpolated parameters, and the smooth distance of shapes is the Chamfer Distance between point clouds sampled from the adjacent interpolated shapess.
The results also show the smoothness of our interpolated shapes, and further demonstrate the quality of the latent space learned by our method.

\subsubsection{Interpolation with Details}

Our approach can also interpolate shapes with details by interpolating the latent codes of the details learned from the VAE. 
For a pair of shapes, we interpolate the details inside the corresponding cuboids by interpolating their latent codes and utilize Algorithm \ref{alg:alg1} to generate meshes. 
The results compared with DSG-Net are shown in Figure \ref{fig:inter}, indicating the efficiency of representing details inside cuboids as three-view drawings. 
The details inside cuboids are also interpolated smoothly with complete and rational details. For example, the backrests of the interpolations firstly have several small holes, then big holes, and finally two sticks as holder. 


\begin{figure}[t]
    \centering
    \includegraphics[width=\linewidth]{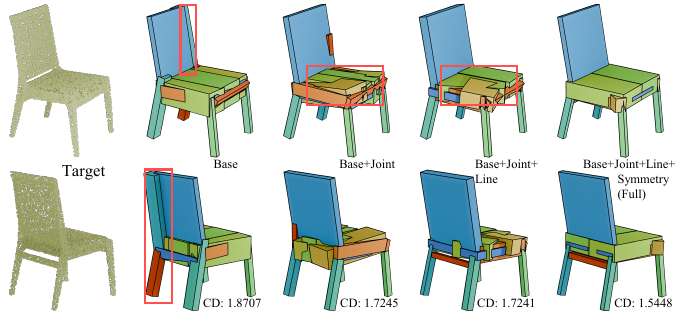}
    \caption{The optimization results of ablated templates on Chair-4 category.}
    \label{fig:ablation}
\end{figure}

\subsection{Ablation Study on Relationship of Cuboids}

We conduct an ablation study on the relationships shown in Figure \ref{fig:design}, including joint, line, and symmetry.
We gradually add these relationships to the template of Chair-4 to restrict the cuboids from a version without any restriction and utilize these templates to optimize shapes from point clouds of Chair-4 category.
As shown in Table \ref{tab:ablation}, the results indicate the necessity of the relationships for the template. 
Figure \ref{fig:ablation} shows an example. Without these relationships, the cuboids are optimized to the wrong position with redundant and chaotic distributions. 
For example one of the legs (green) is optimized to the backrest, and the crossbar (red) of the legs is not represented in the correct position.
The quantitative comparison further demonstrates the necessity of the relationships for the template.

\begin{figure}[t]
    \centering
    \includegraphics[width=0.9\linewidth]{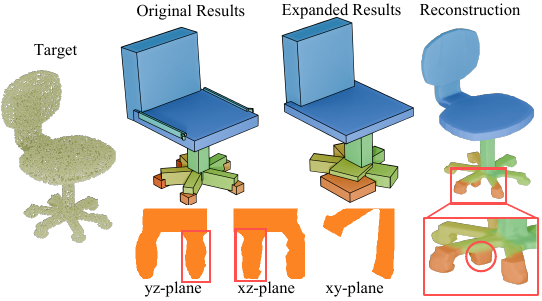}
    \caption{Our predictor trained with Swivel Chair-5 fails to predict the correct shape of a Swivel Chair with 6 legs. The limitation of three-view boundaries leads to wrong detail shapes for complicated shapes inside cuboids.}
    \label{fig:limit}
\end{figure}

\subsection{Limitation and Future Work}
\subsubsection{Limitation}
Our method requires a template for each category, making it hard to process unseen categories. 
As shown in Figure \ref{fig:limit}, the swivel chair has 6 legs, while only 5 legs are defined in the template.  
Thus, the predicted shape misses one leg, leading to wrong results.
We can adjust the nearest cuboid to cover the missed leg and reconstruct the detail of the missed leg. 
However, the wrong shape makes it hard to interpolate and edit the leg. 
Therefore, we need to design a new template for this category. 
To conveniently and rapidly design the template for a new category, we build a GUI system that allows users to modify existing templates or start designing a new template.
The red circle of Figure \ref{fig:limit} also shows the limitation of three-view boundaries. In situations where one cuboid contains a complicated shape, the three-view boundaries may not be able to represent the detail accurately.

\begin{table}[]
\caption{Ablation study of cuboid relationships on Chair-4. ``Base'' indicates the template without restriction. ``Joint'' represents the relationship including ``joint'' and ``restriction'' mentioned in Figure \ref{fig:design}. The results of Chamfer Distance are multiplied by 1000.}
\label{tab:ablation}
\small
\centering
\setlength{\tabcolsep}{8pt}
\begin{tabular}{@{}cccc|l@{}}
\toprule
Base & Joint & Line & Symmetry & CD $\downarrow$ \\ \midrule
\Checkmark &  &  &  & 1.653 \\
\Checkmark & \Checkmark &  &  & 1.643 \\
\Checkmark & \Checkmark & \Checkmark &  & 1.625 \\ \midrule
\Checkmark & \Checkmark & \Checkmark & \Checkmark & \textbf{1.533} (full) \\ \bottomrule
\end{tabular}
\end{table}

\subsubsection{Future Work}
The future work of our method is to automatically design the template for a new category, without utilizing the hierarchical annotation. 
This could be achieved through unsupervised segmenting the point cloud of a new object and generating the bounding box (cuboid) for each part.
After connecting the cuboids with their adjacent areas and finding the reflected part, we can get the template of one object. 
The design for the template of a category needs to combine templates of multiple objects, which requires further research.
Another future work is combining multiple modes to generate new shapes. 
The current proposed approach only generates new shapes with unconditional codes. 
Our generation network is easy to train for generating shapes from conditional codes from images and texts. 
We can render the meshes to images as input images or directly utilize Text2Shape \cite{chen2018text2shape} dataset as the text input to achieve multi-mode generation.

\section{Conclusion}
In this paper, we propose a method to parameterize structure with differentiable templates.
The relationships of cuboids from shared structures for a category are determined with configuration of the template. 
Our method produces the cuboids through a differentiable template from fixed-length parameters representing the shapes.
Detailed shapes inside cuboids are represented with boundaries of three-view drawings of cuboids.
The shape of the objects is recovered from SDFs calculated from parameters and details through the proposed approach.
Benefiting from the proposed representation approach, we employ simple but efficient networks to reconstruct and generate shapes.
We contribute a dataset containing paired point clouds and parameters for training with 20 categories. 
The reconstruction results demonstrate the representation ability of the proposed method.
With a well-structured latent space, our method can generate diverse and rational models.
Adequate qualitative and quantitative comparisons demonstrate the effectiveness and superiority of our method.

\bibliographystyle{IEEEtran}
\bibliography{paper}

\end{document}


\title{Supplementary Materials of ``Parameterize Structure with Differentiable Template for 3D Shape Generation''}

\author{Changfeng Ma, Pengxiao Guo, Shuangyu Yang, Yinuo Chen, Jie Guo, Chongjun Wang, Yanwen Guo, and Wenping Wang,~\IEEEmembership{Fellow,~IEEE}
\IEEEcompsocitemizethanks{\IEEEcompsocthanksitem C. Ma, P. Guo, S. Yang, L.Pu, Y. Li, C. Wang, J. Guo, Y. Guo are with the National Key Lab for Novel Software Technology, Nanjing University, Nanjing 210000, China ( e-mail: changfengma @ smail.nju.edu.com; px guo @ smail.nju.edu.cn; shuangyuyang @ smail.nju.edu.cn; guojie @ nju.edu.cn; chjwang @ nju.edu.cn; ywguo @ nju.edu.cn). 
\IEEEcompsocthanksitem W. Wang is with the Texas A\&M University, United States of America (e-mail: wenping@cs.hku.hk).
\IEEEcompsocthanksitem (Corresponding author: Y. Guo.)
}
}
\markboth{Journal of \LaTeX\ Class Files,~Vol.~14, No.~8, August~2021}%
{Shell \MakeLowercase{\textit{et al.}}: A Sample Article Using IEEEtran.cls for IEEE Journals}


\maketitle

\section{IMPLEMENTATION DETAILS}
\subsection{Calculation from Cuboid Parameters to Transform Matrix}\label{app:cal_vec}
\begin{figure}[t]
    \centering
    \includegraphics[width=\linewidth]{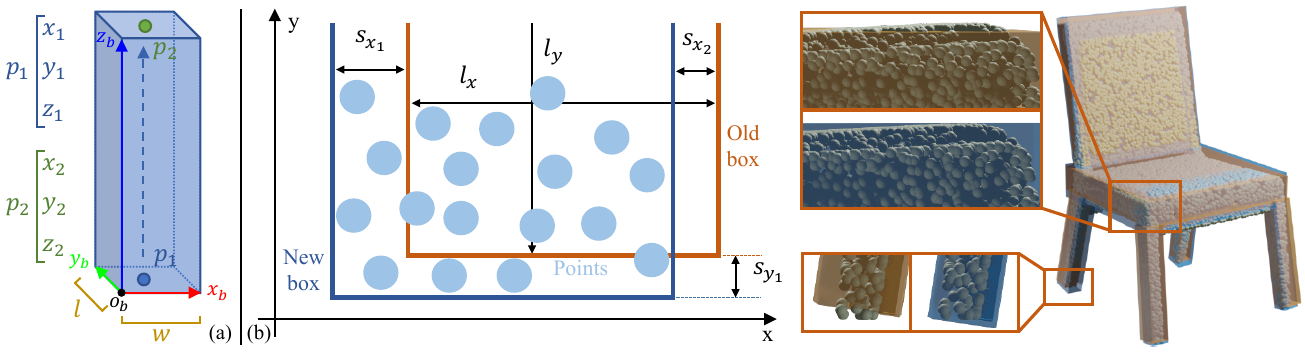}
    \caption{(a):The definition of cuboids utilized in our method. (b): The illustration of expanding cuboids mentioned in the section about limitations.}
    \label{fig:cuboid}
\end{figure}
As shown in Figure \ref{fig:cuboid} (a), let $\bm{p_1}=(x_1,y_1,z_1)$, $\bm{p_2}=(x_2,y_2,z_2)$ and $w$, $l$ denote the cuboid parameters, and $\bm{x_b}, \bm{y_b}, \bm{z_b}, \bm{o_b}$ represent the basis vectors and origin. We calculate $\bm{z_b}$ through:
$$\bm{z_b}=\bm{p_2}-\bm{p_1}.$$
According to Rodrigues' rotation formula, we calculate the rotation matrix $R$ from $\bm{e_z} = (0, 0, 1)$ to $\bm{z_b}$:
\begin{enumerate}
\item Calculate the rotation angle $\theta = \arccos\left(\frac{\bm{z}\cdot\bm{z_b}}{|\bm{z}|\cdot|\bm{z_b}|}\right)$;
\item Calculate the rotation axis vector $\bm{r} = \frac{\bm{z}\times\bm{z_b}}{|\bm{z}\times\bm{z_b}|}$;
\item Calculate the rotation matrix $$R = \cos{\theta}\cdot I + (1-\cos{\theta})\bm{r}^T\bm{r} + \sin{\theta}
    \begin{bmatrix}
        0 & -r_z & r_y \\
        r_z & 0 & -r_x \\
        -r_y & r_x & 0 \\
    \end{bmatrix}
    .$$
\end{enumerate}
Then we have $\bm{x_b}^T = wR\bm{e_x}^T$ and $\bm{y_b}^T = lR\bm{e_y}^T$, where $\bm{e_x} = (1, 0, 0)$ and $\bm{e_y} = (0, 1, 0)$.
The origin $\bm{o_b}$ is calculated through $\bm{o_b} = \bm{p_1} - 0.5 \cdot \bm{x_b} - 0.5 \cdot \bm{y_b}$. And the transforming matrix is 
$$M_b=
    \begin{bmatrix}
        \bm{x_b}^T & \bm{y_b}^T & \bm{z_b}^T & \bm{o_b}^T \\
        0 & 0 & 0 & 1 \\
    \end{bmatrix}
$$
in homogeneous coordinates. Note that all operations are differentiable and can be implemented in \textit{PyTorch}.

Sampling the point cloud from a cuboid can be achieved:
\begin{enumerate}
\item Sampling the point cloud $\mathcal{P}_u$ inside the unit cube or on the surface of the unit cube;
\item Transform $\mathcal{P}_u$ utilizing $M_b$: $\mathcal{P}_b = M_b\mathcal{P}_u$.
\end{enumerate}
This process is also differentiable. We fuse all the point clouds sampled from cuboids as the sampling results of the structure.

\subsection{Detailed explanation from definition to differentiable computation graph}

\begin{figure*}[t]
    \centering
    \includegraphics[width=0.9\linewidth]{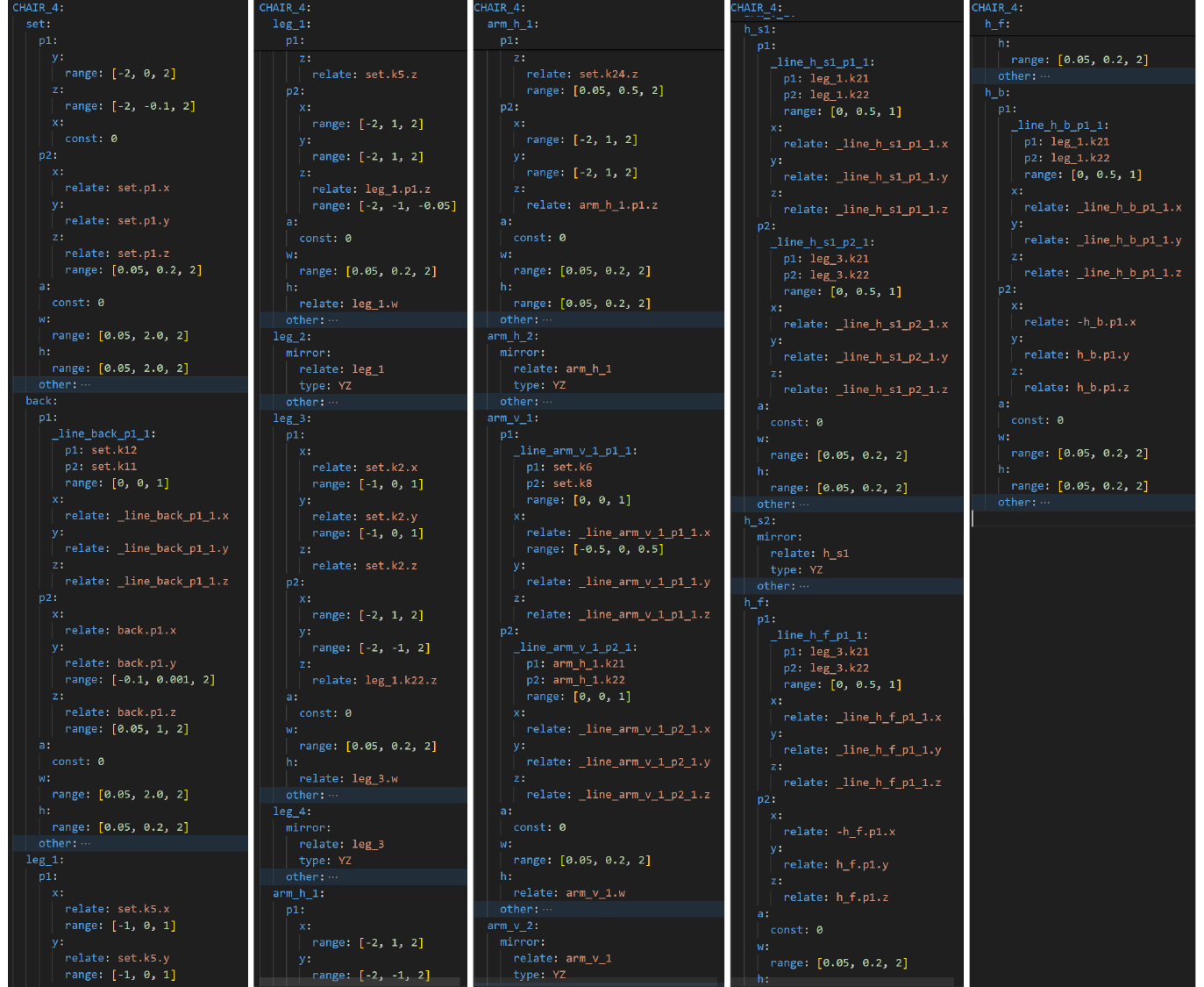}
    \caption{The configuration of the template for Chair-4.}
    \label{fig:config}
\end{figure*}

\begin{table*}[]
\centering
\caption{Examples from codes to arguments of Equation 1. $N_{para}$ indicates the number of the parameter control $b_i$. $\#_{X}$ represents the index of cuboid $X$.}
\label{tab:code2equ}
\begin{tabular}{l|c|c|cccc|cccccc|c}
\toprule
Code & $c_i$ & $r_i$ & $s_{i1}$ & $j_1$ & $k_1$ & $\bm{e_1}$ & $s_{i2}$ & $j_2$ & $k_2$ & $j_3$ & $k_3$ & $\bm{e_2}$ & $N_{para}$ \\ \midrule
const: 0.5 & 0.5 & 0 & 0 &  &  &  & 0 &  &  &  &  &  & 0  \\ \midrule
range: [-1, 0, 1] & -1 & 2 & 0 &  &  &  & 0 &  &  &  &  &  & 1  \\ \midrule
relate: -C.k4.x & 0 & 0 & -1 & $\#_C$ & 4 & x & 0 &  &  &  &  &  & 0  \\ \midrule
\makecell[l]{range: [-1, 0, 1] \\ const: 0.5} & -0.5 & 2 & 0 &  &  &  & 0 &  &  &  &  &  & 1  \\ \midrule
\makecell[l]{range: [-1, 0, 1] \\ relate: B.k9.y} & -1 & 2 & 1 & $\#_B$ & 9 & y & 0 &  &  &  &  &  & 1  \\ \midrule
\makecell[l]{line: p1: D.k2, p2: E.k6 \\ relate: line.z} & 0 & 0 & 0 &  &  &  & 1 & $\#_D$ & 2 & $\#_E$ & 6 & z & 1 \\ \bottomrule
\end{tabular}
\end{table*}

In practice, we utilize \textit{json} files to store the configuration of templates in the form of codes. 
Figure \ref{fig:config} shows the configuration of the template for Chair-4.
Table \ref{tab:code2equ} shows several rules of how to transform the code to the arguments of Equation 1 for building the differentiable computation graph.
Here, $a$, $b$, $c$ in ``range:[$a$, $b$, $c$]'' represent the minimum value, the default value, and the maximum value of the parameter $a_{i1}$, separately.
$\bm{K}_{jk}$ is obtained by:
\begin{enumerate}
\item Calculate the transform matrix of the $j$-th cuboid $M_j$;
\item Get the coordinate $\bm{p}_k$ of the $k$-th key point on unit cube;
\item Transform $\bm{p}_k$ utilizing $M_j$: $\bm{K}_{jk} = M_j\bm{p}_k$.
 \end{enumerate}
Most $b_i$ do not have corresponding parameters $a_{i1}$ and $a_{i2}$, and $a_{i1}$ and $a_{i2}$ do not occur at the same time.
$b_i$ can also be related with $b_j$ ($j < i$), such as the relationship (3) of the Figure 3 in the paper. This can be achieved by directly copy the value of $b_j$ to $b_i$. 
We build the differentiable computation graphs according to Table \ref{tab:code2equ} from definition \textit{json} files.

\subsection{Automatically Design Template from the Hierarchical Annotation}

We also propose an approach to automatically design the template from the hierarchical annotation of an object.
The hierarchical annotation is obtained from PartNet \cite{mo2019partnet} with the same preprocessing of StructureNet \cite{StructureNet}.
This automatic process is achieved by:
\begin{enumerate}
    \item Change the definitions of the cuboids from transform matrix to ``stick'' definition:
    \begin{enumerate}
        \item Find the longest axis of the cuboid;
        \item Get two control points and the size of the cuboid.
    \end{enumerate}
    \item Check if each cuboid is symmetrical to itself:
    \begin{enumerate}
        \item Normlize the cuboid to the center;
        \item Reflect the cuboid across YZ, XZ, and XY plane;
        \item Calculate the distance between the reflection and the cuboid;
        \item If the distance is lower than the threshold, the cuboid is symmetrical to itself;
        \item Use relationship (3) of Figure 3 in the paper to describe this cuboid.
    \end{enumerate}
    \item Check if each cuboid is symmetrical to others, similar to the self-symmetry check.
    \item Find joints for each cuboid:
    \begin{enumerate}
        \item Take one of the control points of a cuboid;
        \item Calculate the distances between the control point and all the key points of other cuboids;
        \item Find the closest key point;
        \item If the distance is lower than the threshold, there is a joint;
        \item Use relationship (1) of Figure 3 in the paper to describe this control point.
    \end{enumerate}
\end{enumerate}
The cuboids are named with the semantic and instance information offered by the hierarchical annotation.

\subsection{Detail of Reconstruction from point cloud}

The encoder and decoder of our reconstruction network are PointNet++ \cite{qi2017pointnet++} and MLPs, respectively. The encoder takes point clouds with 2048 points as input and outputs a global feature with 1024 channels. Then the decoder takes the global feature as input and outputs the parameters. The channels for each layer in MLPs are 1024, 512, 256, and $N_a$. Here, $N_a$ denotes the parameter number.
We employ the MSE between predicted paremters and ground truth as the loss during training. 

\begin{figure}[t]
    \centering
    \includegraphics[width=\linewidth]{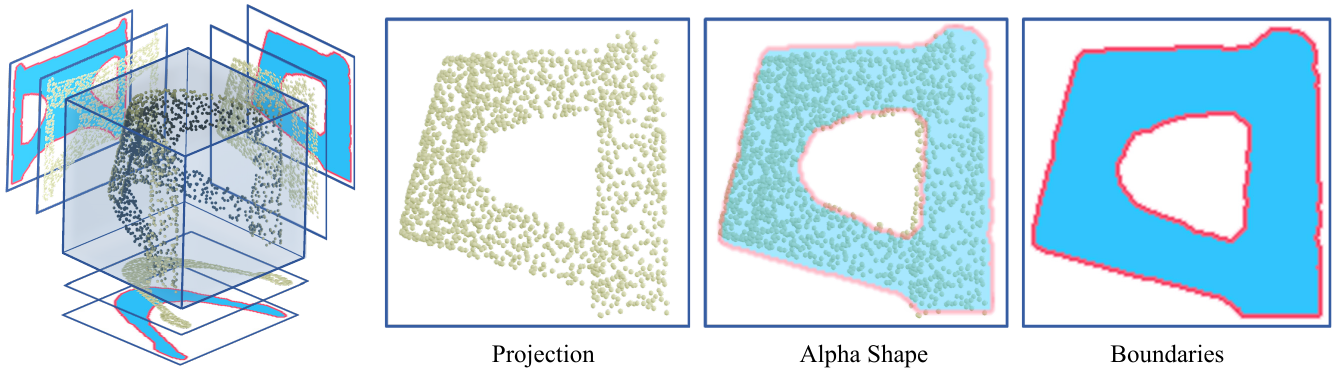}
    \caption{The process of extracting the boundaries of the three-view detail from point cloud inside cuboids.}
    \label{fig:recon}
\end{figure}

The detailed process of the whole reconstruction approach is:
\begin{enumerate}
    \item Reconstruct the structure from point cloud:
    \begin{enumerate}
        \item Predict the parameter from the point cloud utilizing the reconstruction network;
        \item Produce the cuboids representing structure through the differentiable template;
    \end{enumerate}
    \item Reconstruct the details inside cuboids (as shown in Figure \ref{fig:recon}):
    \begin{enumerate}
        \item Split point clouds inside each cuboid;
        \item Normalize the point cloud into the unit cube;
        \item Project the point cloud to the three-view drawing;
        \item Apply Alpha Shape \cite{alphashape} Algorithm on 2D projection;
        \item Store the boundaries as the detail inside the cuboid.
    \end{enumerate}
    \item Utilize Algorithm 1 in the paper to recover the mesh.
\end{enumerate}

\subsection{Detail of Generation}

\subsubsection{Structure Generation}

The encoder and decoder of VAE\cite{VAE_Kingma2013AutoEncodingVB} are MLPs. The encoder takes parameters as input and outputs mean code $\mu$ and variance code $\sigma$ with 32 channels. The channels for each layer in MLPs of the encoder are $N_{para}$, 512, 256, 128, 128, 128, and 32. The channels for each layer in MLPs of the decoder are 32, 128, 128, 256, 256, 256, and $N_{para}$. $N_{para}$ indicates the parameter length.

We random sample the code $c_{norm}$ from the standard normal distribution and get the latent code $c_{real} = \sigma * c_{norm} + \mu$ as the real input of the discriminator.
The discriminator predicts the probability that the input is real. 
We train VAE with losses including:
\begin{itemize}
    \item MSE loss for parameter reconstruction;
    \item KL divergence loss.
\end{itemize}

\subsubsection{Detail Generation}

We draw the boundaries into binary images with 128$\times$ 128 resolution, and fill the inside with ``1'' (white indicates inside, black indicates outside).
We then resize the 2D image into a 1D vector as the input of VAE. Networks with the same architecture  are employed. 
The channel of the latent code utilized for detail is 128.
The channels for each layer in MLPs of the encoder are 128*128, 1024, 1024, 512, 256, 256, and 128. The channels for each layer in MLPs of the decoder are 128, 256, 256, 512, 1024, 1024, and 128*128. 
The reconstruction loss of VAE for detail generation is also the MES loss.

The detailed process of the whole generation approach is:
\begin{enumerate}
    \item Generate a parameter.
    \item Produce the cuboids representing shape through the template according to the parameter.
    \item Generate detail for each cuboid:
    \begin{enumerate}
        \item Generate the three-view binary images for the cuboid;
        \item Find the edge curves from binary images and store the edge curves as the boundaries of the three-view drawing as the detail;
        \item If the cuboid is symmetric to another cuboid, just reflect its three-view boundaries.
    \end{enumerate}
    \item Utilize Algorithm 1 in the paper to recover the mesh.
\end{enumerate}

\begin{figure}[t]
    \centering
    \includegraphics[width=\linewidth]{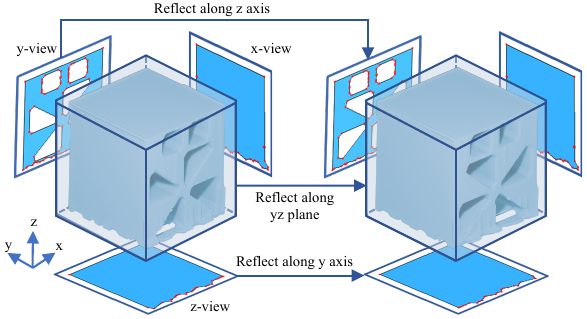}
    \caption{We reflect a detail by reflecting the boundaries of three-view images.}
    \label{fig:reflect}
\end{figure}

Reflecting detail is easy to achieve by reflecting the vertices of boundaries. As shown in Figure \ref{fig:reflect}, to reflect a detail along the yz-plane, we just need to reflect the y-view detail along the z-axis and reflect the z-view along the y-axis.

\subsection{Detail of Interpolation}

The detailed process of the whole interpolation approach is:
\begin{enumerate}
    \item Get the latent codes of two objects by putting their parameters into the encoder of the VAE.
    \item Interpolate the latent codes and get the parameter by putting interpolated latent codes into the decoder of the VAE. Or directly interpolate the parameter.
    \item Produce the cuboids representing structure through the differentiable computation graph.
    \item Interpolate the details inside cuboids:
    \begin{enumerate}
        \item Draw the boundaries of the three-view on binary images;
        \item Put the images into the encoder of the VAE for details and get the latent code; 
        \item Interpolate the latent codes and get the interpolated images utilizing the decoder of the VAE for details;
        \item Find the edge curves from binary images and store the edge curves as the boundaries of the three-view drawing as the detail.
    \end{enumerate}
    \item Utilize Algorithm 1 in the paper to recover the mesh.
\end{enumerate}

\subsection{Implementation Details}

\begin{figure}[t]
    \centering
    \includegraphics[width=\linewidth]{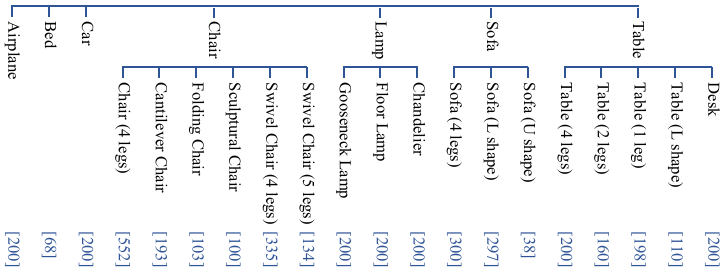}
    \caption{Visualization of data numbers of each category. We subdivide Chair, Lamp, Table, and Sofa into more categories.}
    \label{fig:dataset}
\end{figure}

We implement our method with PyTorch \cite{nips33pytorch}. We will make the code and dataset publicly available.
We train our network on NVIDIA 2080Ti utilizing Adam Optimizer with the learning rate $1\times 10^{-4}$ and the batch size $4$.
The training for reconstruction network and VAE takes nearly 20 and 100 minutes, separately. 
Algorithm 1 in the paper, which calculates SDFs given parameters and details, is implemented with a multi-thread program and takes nearly 10-20 seconds to recover represented shapes into meshes.

We select nearly 4,000 models from the ShapeNet dataset \cite{chang2015shapenet}, and classify the models into 20 categories. 
Figure \ref{fig:dataset} shows the names and the data numbers of 20 categories.
Each data of our dataset contains the point cloud, the annotated parameter of the object and details inside cuboids.
The point cloud is sampled from mesh of ShapeNet.
The parameter is annotated according to the point cloud utilizing our annotated system. The parameter is first optimized by our method and then manually annotated. Annotating an object usually takes 0.5-1 minutes. 
The details inside cuboids is automatically reconstructed by our reconstruction method based on the annotated parameter.

\subsection{Expand cuboids to Fit Missed Points}

Given a point cloud $\mathcal{P}\in \mathbb{M}_{n\times 3}$ and the cuboids generated by predicted parameters, we first label the points inside cuboids (points can belong to multiple cuboids). Then for an unlabeled point that is outside the cuboids, we calculate its distance to each cuboid and label it with the information of the nearest cuboid. Then for each cuboid, we adjust its width, length, height and origin point according to the boundaries of the points belonging to it, as shown in Figure \ref{fig:cuboid} (b).

We define the distance $d$ from a point $\bm{p}$ to a cuboid $B$ as the maximum vertical distance from $\bm{p}$ to the 6 faces.
To calculate the distance, we first inverse transform $p$:
$$\bm{p'}^T = M_b^{-1}\bm{p}^T.$$
Then we calculate the distance through:
$$\bm{d_1}=abs(\bm{p'})\cdot(|\bm{b_x}|, |\bm{b_y}|, |\bm{b_z}|)=(d_{1x}, d_{1y}, d_{1z}),$$
$$\bm{d_2}=abs(\bm{p'}-1)\cdot(|\bm{b_x}|, |\bm{b_y}|, |\bm{b_z}|)=(d_{2x}, d_{2y}, d_{2z}),$$
$$\bm{d}=\max(d_{1x}, d_{1y}, d_{1z}, d_{2x}, d_{2y}, d_{2z}).$$

For a cuboid $B$, we calculate the rescale parameters $s_{x_1}=0-x_{min}$, $s_{x_2}=x_{min}-1$, $s_{y_1}=0-y_{min}$, $s_{y_2}=y_{min}-1$, $s_{z_1}=0-z_{min}$, $s_{z_2}=z_{min}-1$ for 6 faces, where $x_{min}$ and $x_{max}$ indicate the minimum and maximum x-coordinate of the points belong to $B$. Then new basis vectors and the new origin of box $B$ are:
$$\bm{o_b}' = \bm{o_b} - s_{x_1}\bm{x_b} - s_{y_1}\bm{y_b} - s_{z_1}\bm{z_b},$$
$$\bm{x_b}' = (1+s_{x_1}+s_{x_2})\bm{x_b},$$
$$\bm{y_b}' = (1+s_{y_1}+s_{y_2})\bm{y_b},$$
$$\bm{z_b}' = (1+s_{z_1}+s_{z_2})\bm{z_b}.$$
The ``Expanded Results'' in the Limitation Section of the paper is acquired through this approach.

\subsection{System for Annotation, Designing, and Applications}


\begin{figure*}[t]
    \centering
    \includegraphics[width=\linewidth]{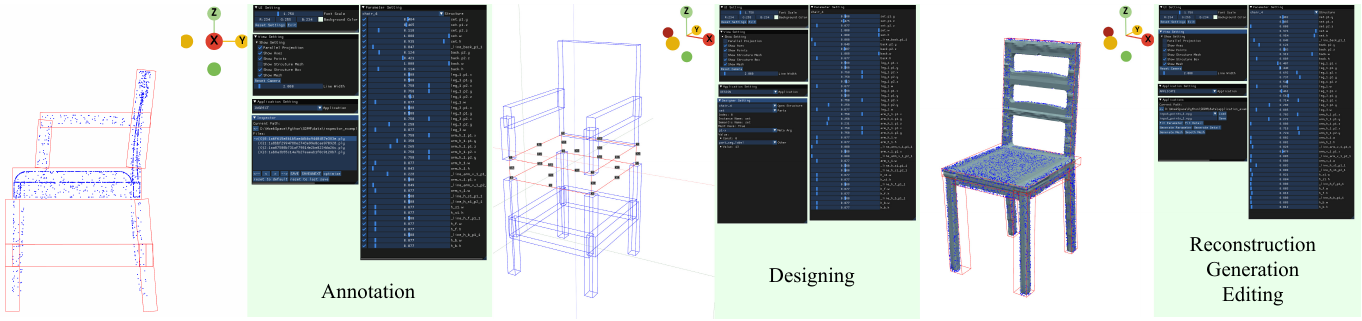}
    \caption{The GUI of our system for annotation, designing, and applications.}
    \label{fig:sys}
\end{figure*}

We also build a system for users to conveniently parameterize new categories and annotate objects, to facilitate the use of our method. The GUI of our system for annotation, designing, and applications are shown in Figure \ref{fig:sys}.
With our system, users can efficiently annotate the shape structures for point clouds, design a new definition for a category, and apply reconstruction, generation, and structures edition.
The demonstration of our system is shown in the video of our supplementary materials.
We will make the code for training and the code of system publicly available.

\section{More Results and Experiments}

\subsection{Data Augmentation for Structure Reconstruction}
As depicted in Figure \ref{fig:dataset}, some categories have limited data, which is not conducive to network training.
We replace point clouds of object parts for data augmentation to improve the performance of our parameter prediction network. 
The MSE between predicted parameters and ground truth parameters on the test set of different categories with and without data augmentation are shown in Table \ref{tab:aug_com}.   
Data augmentation enhances the performance of our network in categories with limited sizes, such as Sofa-U, Bed, and Sculptural Chair. This improves the robustness of our network with limited data, reducing the reliance on large datasets.

\begin{table}[]
\centering
\caption{The MSE of parameters predicted by predictors that are trained with and without data augmentation. Data augmentation promotes the network when the dataset is limited.}
\label{tab:aug_com}
\begin{tabular}{@{}r|cc|c@{}}
\toprule
\multicolumn{1}{l|}{\textbf{}} & w/o aug. & +aug.  & size \\ \midrule
Sofa - U                       & 0.0156   & 0.0122 & 38   \\
Bed                            & 0.0190   & 0.0174 & 68   \\
Sculptural Chair               & 0.0135   & 0.0087 & 100  \\
Cantilever Chair               & 0.0055   & 0.0044 & 193  \\
Airplane                       & 0.0009   & 0.0007 & 200  \\
Gooseneck lamp                 & 0.0068   & 0.0068 & 200  \\ \bottomrule
\end{tabular}
\end{table}

\subsection{All Templates}

All the 20 categories in our dataset and corresponding descriptions are in Table \ref{tab:cate_desc}. 
We used these criteria to categorize the data.
The templates and parameter numbers of all definitions can be viewed in Figure \ref{fig:app_allcate}.



\subsection{K Nearest Neighbors of Generated Shapes}

Figure \ref{fig:nn} shows the 3 nearest neighbors of the generated shapes on the distance of geometry and parameter. 
The results show that our method can generate diverse shapes without copying the training data directly.
Meanwhile, the similarity of the nearest neighbors about geometry distance and parameter distance shows that the parameter distance is meaningful enough to represent the geometry distance between shapes.
This also shows our representation utilizing parameters is reasonable.

\subsection{Results of Interpolating Parameters and More Interpolation Results}

Figure \ref{fig:inter2mothed} shows the interpolated shapes of two interpolation approaches. 
The similarity of the interpolation process on parameters and codes also verifies the rationality of our method.
Since parameters directly control the shapes, the interpolated results are smoother.

Figure \ref{fig:supp_inter} shows more interpolated shapes of our method on Chair and Table categories.

\subsection{Results of Editing Shapes}

\begin{figure}[t]
    \centering
    \includegraphics[width=\linewidth]{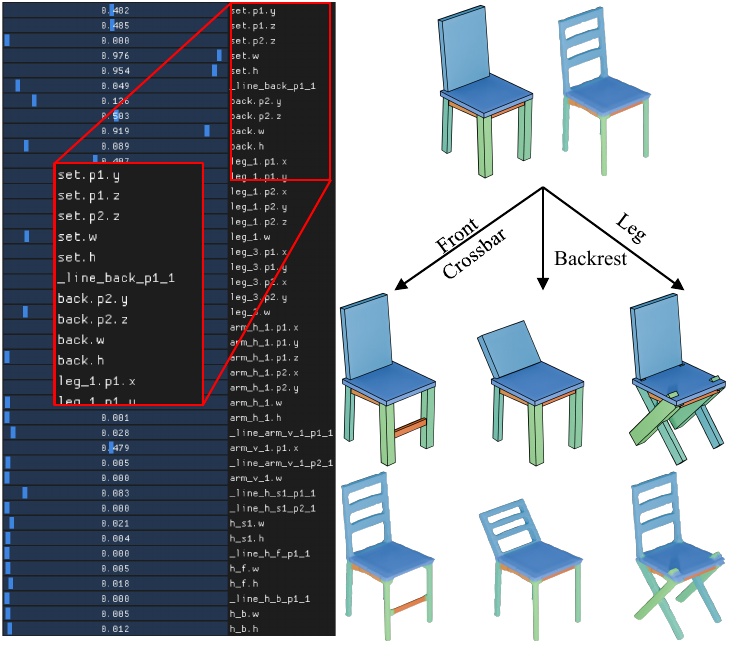}
    \caption{The parameters of our method have names with practical meanings. The right part shows the results of editing the front crossbar, the backrest, and the legs of the chair on the top.}
    \label{fig:edit}
\end{figure}

Traditional shape editing methods \cite{ProbReason, SemanticEditing} employ  complicated rules to edit the shape object for several certain categories. 
Recent deep-learning-based approaches \cite{hao2020dualsdf, hertz2022spaghetti} utilize different neural primitives including sphere and Gaussian to represent objects and recover them to meshes through decoders. The object can be edited by moving the primitives.
Previous works GRASS \cite{GRASS} and StrucutureNet \cite{StructureNet} do not restrict the cuboids. Thus, each cuboid of structures generated from these two methods is defined with 9 parameters as a transform matrix. 
It is not intuitive for users to adjust the transform matrix for editing the cuboid. 
After editing, these two methods also can not guarantee the rationality of the edited structure. 
ShapeAssembly \cite{ShapeAssembly} utilizes programs to represent structures, which requires users to learn the grammar to edit the structure. 
ShapeMOD \cite{ShapeMOD} and ShapeCoder \cite{ShapeCoder} abstract structural patterns from programs, that focus on decreasing the number of the parameter. Their abstract structural patterns are learned from networks. Thus, it may be hard for users to understand the mapping from abstracted parameters to cuboids. 
The parameters of our method directly control the actual part of the cuboids, which benefits from our stick-like definition of the cuboid.
This also achieves precise control of the shape. 
As shown in Figure \ref{fig:edit}, these parameters are actually the coordinates of the control points of the cuboid, which is intuitive.
After the shapes are edited, we can still obtain the mesh with the original details by Algorithm 1 in the paper.
Figure \ref{fig:edit} shows several edited shapes.

\subsection{More Results of Reconstruction and Generation}

Figure \ref{fig:supp_recon} shows more reconstructed shapes of our method from point clouds.

Figure \ref{fig:supp_gen} shows more generated shapes of our method.

\bibliographystyle{IEEEtran}
\bibliography{paper}

\begin{table*}[]
\small
\centering
\caption{Categories in our dataset and corresponding descriptions. More intuitive presentation is in Figure \ref{fig:app_allcate}.}
\label{tab:cate_desc}
\begin{tabular}{@{}rl@{}}
\toprule
Category & Description                                                                                             \\ \midrule \midrule
Airplane          & Typical airliner, usually consisting of a fuselage, wings, tail, engines and tires.                     \\ \midrule
Bed               & Common bed, with or without bedside tables.                                                             \\ \midrule
Car               & Common car with four tires, including sedans, vans, sports cars and so on.                              \\ \midrule
Chair-4         & Common four-legged chair with or without armrests / crossbars between legs.                             \\ \midrule
Cantilever Chair  & Chair with two symmetrical L-shape legs as supporting structures, there may be a crossbar between legs. \\ \midrule
Folding Chair     & Common folding chair with the front legs and back legs in an A-shape or X-shape.                                  \\ \midrule
Sculptural Chair  & Similar to the Cantilever Chair, but with two legs fused as a whole.                                          \\ \midrule
Swivel Chair-5  & Swivel chair with five wheels, with or without arms.                                                    \\ \midrule
Swivel Chair-4  & Swivel chair with four wheels, with or without arms.                                                    \\ \midrule
Gooseneck Lamp    & Lamp with bending supporting structures, usually consisting of several sections connected in sequence.  \\ \midrule
Floor Lamp        & Lamp standing vertically, usually with luminescent part at the top center.                        \\ \midrule
Chandelier        & Light suspended from the ceiling, typically with luminescent part at the bottom center.           \\ \midrule
Sofa-4          & Common single sofa with four legs.                                                                      \\ \midrule
Sofa-L          & Corner sofa, `L' means the shape is like the capital letter L.                                          \\ \midrule
Sofa-U          & Similar to the shape of Sofa - L, but the shape is like the capital letter U.                     \\ \midrule
Table-4         & Common table with four legs for support, with or without crossbars between legs.                        \\ \midrule
Table-2         & Table with only two legs on the side.                                                            \\ \midrule
Table-1         & Table with only one leg in the center as supporting structures.                                        \\ \midrule
Table-L         & Corner table, `L' means the shape is like the capital letter L.                                         \\ \midrule
Desk              & Similar to Table-2, but the two legs have complex structures, with several drawers for example.       \\ \bottomrule
\end{tabular}
\end{table*}

\begin{figure*}[t]
    \centering
    \includegraphics[width=0.9\linewidth]{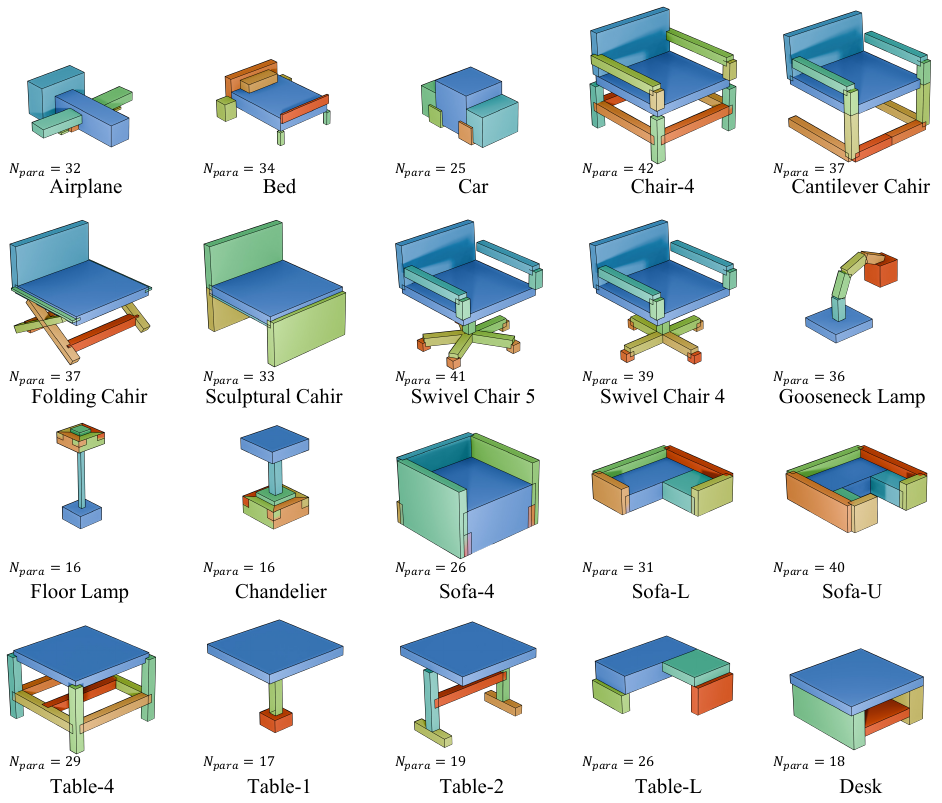}
    \caption{Templates of the 20 categories in our dataset, the name of each category is shown under the figure while the number in the figure represents the number of parameters of each category.}
    \label{fig:app_allcate}
\end{figure*}

\begin{figure*}[t]
    \centering
    \includegraphics[width=0.9\linewidth]{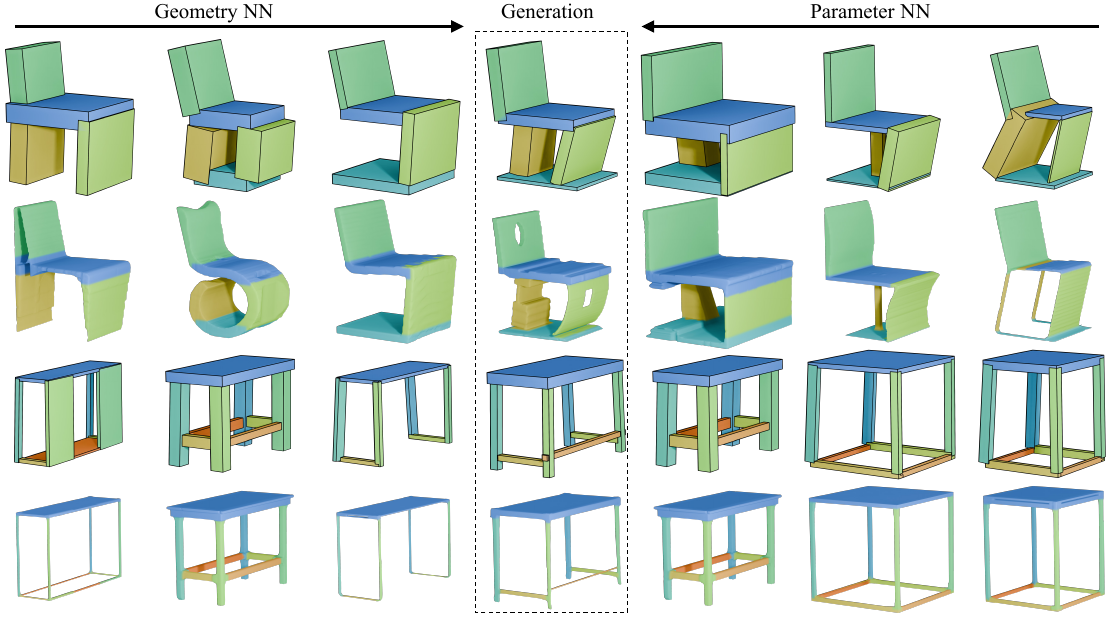}
    \caption{The 3 nearest neighbors of generated shapes. ``Geometry NN'' represents the nearest neighbors on the Chamfer Distance of geometries. ``Parameter NN'' represents the nearest neighbors on the MSE of parameters. The model has a smaller distance when closer to the center.}
    \label{fig:nn}
\end{figure*}

\begin{figure*}[t]
    \centering
    \includegraphics[width=\linewidth]{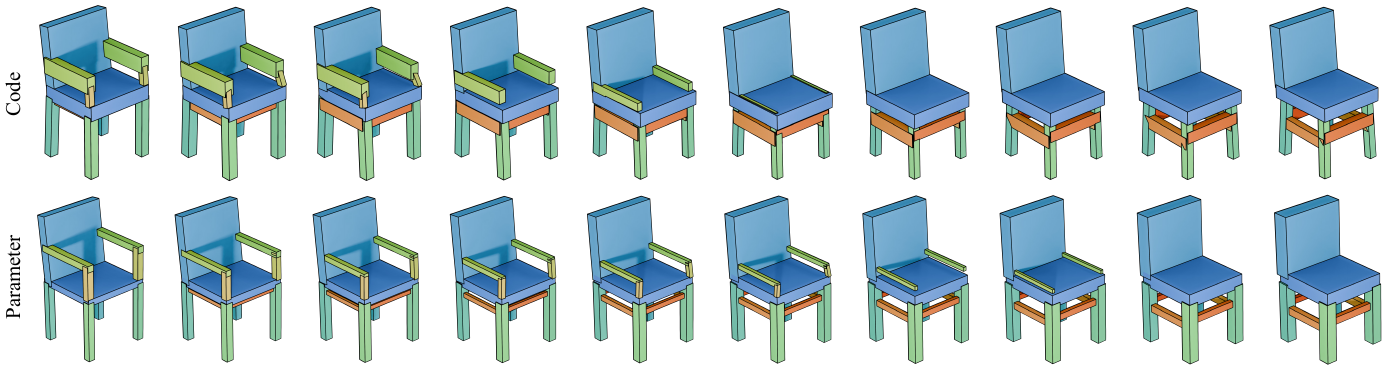}
    \caption{The interpolated results of two approaches. The first row is interpolated results utilizing latent codes. The second row is interpolated results utilizing parameters.}
    \label{fig:inter2mothed}
\end{figure*}

\begin{figure*}[t]
    \centering
    \includegraphics[width=\linewidth]{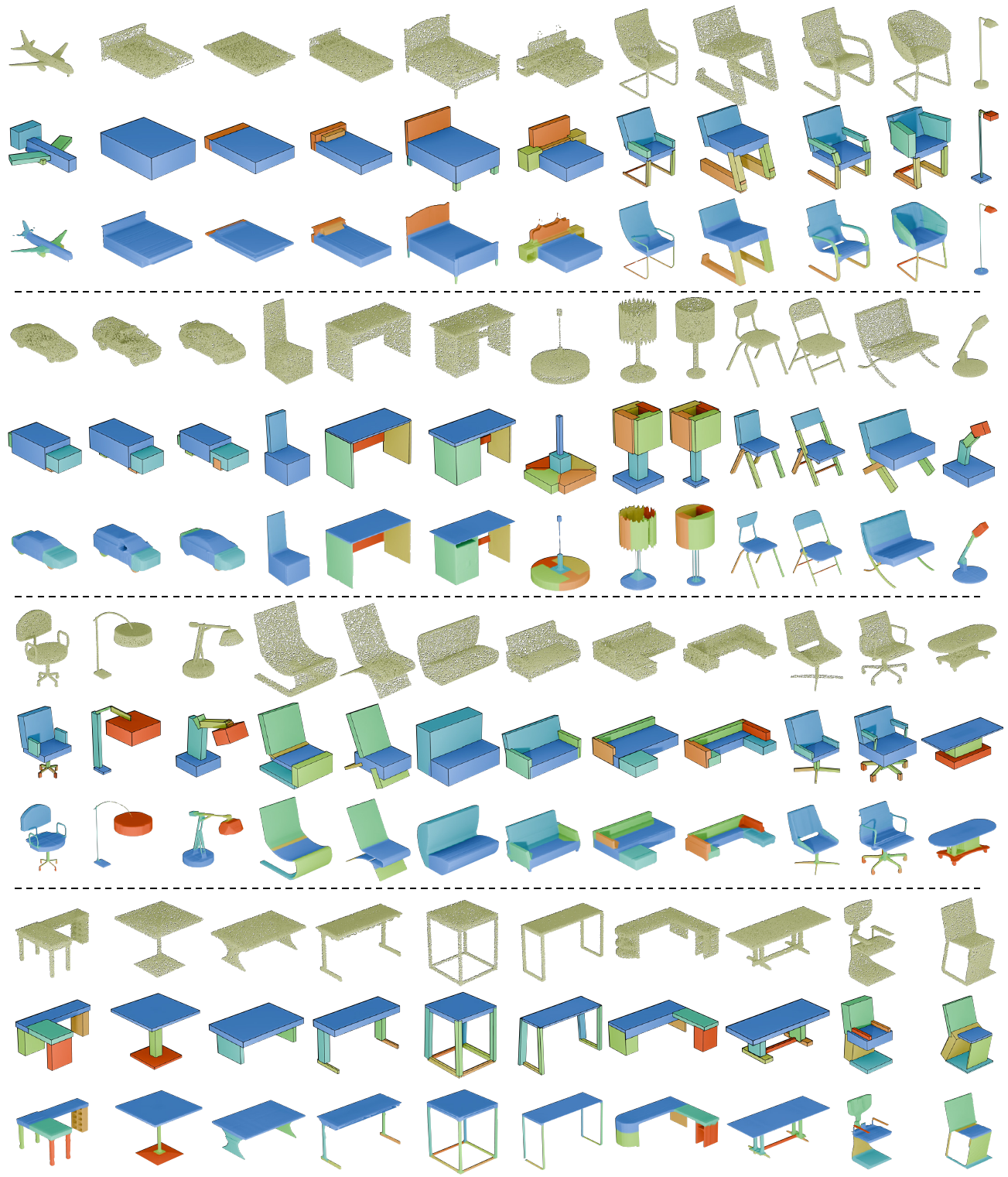}
    \caption{More reconstructed shapes of our method from point clouds.}
    \label{fig:supp_recon}
\end{figure*}

\begin{figure*}[t]
    \centering
    \includegraphics[width=\linewidth]{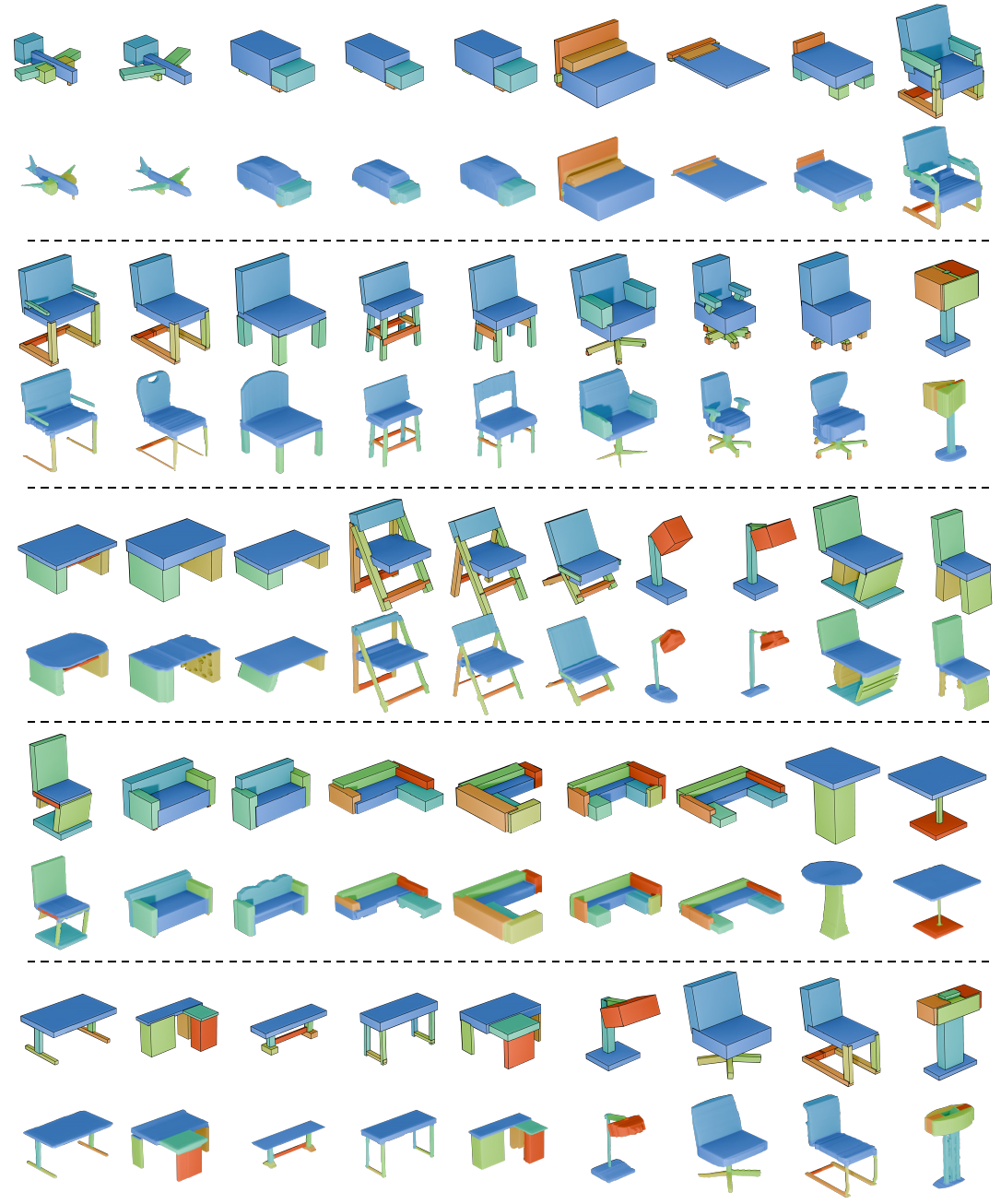}
    \caption{More generated shapes of our method.}
    \label{fig:supp_gen}
\end{figure*}

\begin{figure*}[t]
    \centering
    \includegraphics[width=\linewidth]{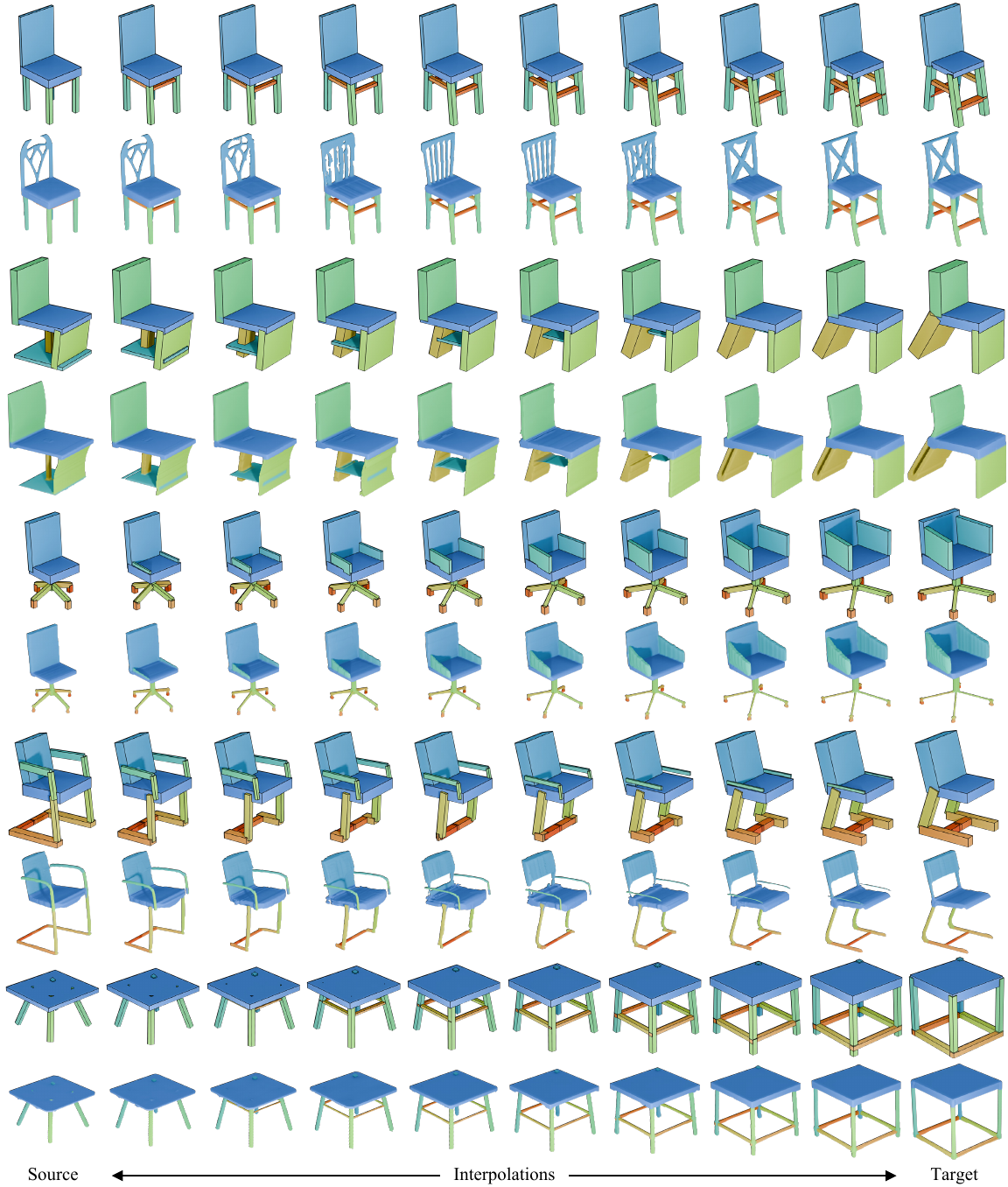}
    \caption{More interpolated shapes of our method on Chair and Table categories.}
    \label{fig:supp_inter}
\end{figure*}